\documentclass[sn-mathphys]{sn-jnl}
\usepackage{subfigure}
\usepackage{caption}

\jyear{2023}%

\theoremstyle{thmstyleone}%
\theoremstyle{thmstyletwo}%

\theoremstyle{thmstylethree}%

\newcommand\XP[1]{\textcolor{black}{#1}}

\begin{document}

\title[Wang et al, Towards Practical Multi-Robot Hybrid Tasks Allocation]{Towards Practical Multi-Robot Hybrid Tasks Allocation for Autonomous Cleaning}

\author[1]{\fnm{Yabin} \sur{Wang}}\email{iamwangyabin@stu.xjtu.edu.cn}

\author*[1,2]{\fnm{Xiaopeng} \sur{Hong}}\email{hongxiaopeng@ieee.org}

\author[3]{\fnm{Zhiheng} \sur{Ma}}\email{zh.ma@siat.ac.cn}

\author[4]{\fnm{Tiedong} \sur{Ma}}\email{tdma@cqu.edu.cn}

\author[5]{\fnm{Baoxing} \sur{Qin}}\email{baoxing@gs-robot.com}

\author[1]{\fnm{Zhou} \sur{Su}}\email{zhousu@ieee.org}

\affil*[1]{\orgdiv{The school of Cyber Science and Engineering}, \orgname{Xi'an Jiaotong University}, \orgaddress{\city{Xi'an}, \country{China}}}

\affil[2]{\orgdiv{Faculty of Computing}, \orgname{Harbin Institute of Technology},  \orgaddress{\city{Harbin}, \country{China}}}

\affil[3]{\orgdiv{Shenzhen Institute of Advanced Technology}, \orgname{Chinese Academy of Science}, \orgaddress{\city{Shenzhen}, \country{China}}}

\affil[4]{\orgname{School of Automation, Chongqing University}, \orgaddress{\city{Chongqing}, \country{China}}}

\affil[5]{\orgname{Gaussian Robotics}, \orgaddress{\city{Shenzhen}, \country{China}}}


\abstract{Task allocation plays a vital role in multi-robot autonomous cleaning systems, where multiple robots work together to clean a large area. However, most current studies mainly focus on deterministic, single-task allocation for cleaning robots, without considering hybrid tasks in uncertain working environments. Moreover, there is a lack of datasets and benchmarks for relevant research. In this paper, to address these problems, we formulate  multi-robot hybrid-task allocation under the uncertain cleaning environment as a robust optimization problem. Firstly, we propose a novel robust mixed-integer linear programming model with practical constraints including the task order constraint for different tasks and the ability constraints of hybrid robots. Secondly, we establish a dataset of \emph{100} instances made from floor plans, each of which has 2D manually-labeled images and a 3D model. Thirdly, we provide comprehensive results on the collected dataset using three traditional optimization approaches and a deep reinforcement learning-based solver. The evaluation results show that our solution meets the needs of multi-robot cleaning task allocation and the robust solver can protect the system from worst-case scenarios with little additional cost.
The benchmark will be available at \href{https://github.com/iamwangyabin/Multi-robot-Cleaning-Task-Allocation}{https://github.com/iamwangyabin/Multi-robot-Cleaning-Task-Allocation}.}




\maketitle

\section{Introduction}\label{sec1}

Cleaning robots have been widely applied in various environments such as libraries, airports, playgrounds, and warehouses. 
{One important research topic related to cleaning robots is task allocation, which aims to effectively assign tasks to robots to minimize  energy consumption and complement time.}

According to different application scenarios, the task allocation methods for cleaning robots can be briefly divided into two {categories}: {low-level planning and high-level scheduling.} The low-level planning is \emph{a.k.a.} the complete coverage path planning~\cite{cabreira2019survey, almadhoun2019survey, lakshmanan2020complete}, which provides  specific paths for cleaners to traverse every accessible area in a given cleaning region. Nonetheless, the computational complexity of low-level task allocation strategy increases drastically when the dimension of the problem grows \cite{almadhoun2019survey, 10.5555/3463952.3463974}. Moreover, the solution may easily fail into local optima \cite{shang2020co}. \XP{In contrast,} high-level {scheduling}~\cite{huang2018multiple, 7361161, ahmadi2006multi, jeon2016multiple, jiang2019group}, which firstly partitions the whole cleaning space into several regions and then assigns robots to clean these regions, is more suitable for multi-robot cleaning in large public spaces such as libraries, airports, playgrounds, and warehouses.
In this paper, we mainly focus on  high-level task allocation for clearning robots. 

{The study of high-level task allocation has limitations that need to be addressed.}
Firstly, existing approaches for task allocation are usually designed for deterministic environments, {where the next state of the environment is SOLELY determined by its current state and the actions performed by the agents. Accordingly, models for solving the problem are also deterministic, which will always produce the same output from a given initial state.} However, there is a great deal of uncertainty involved in actual scenarios, such as the movement of objects and the randomness of garbage generation. Thus uncertainty shall be considered in modelling future states of the environments. The aforementioned deterministic methods cannot 
model such uncertainty. Thus approaches for multi-cleaning robot task allocation under uncertainty are highly desired.

Secondly, most existing task allocation {studies} only consider {the \emph{single-task allocation} problem, where robots accomplish a single type of task}~\cite{ahmadi2006multi,7361161, jeon2016multiple, wang2021hybrid}.
However, in practice, it is normal that different places need different cleaning tasks and a specific kind of robots may only be able to accomplish a part of the tasks. Thus \emph{multi-robot hybrid cleaning task allocation} needs to be urgently addressed.
What's more, in many cases, there exists dependence between tasks. For example, vacuuming throughout the house usually shall be performed before mopping and disinfecting.
{Nonetheless}, few existing studies jointly consider hybrid cleaning tasks and task order constraint.

Thirdly, there are \XP{few} public benchmarks specifically designed for multi-robot cleaning task allocation. 
Current studies usually use small-size randomly generated data to test approaches and the performance on real large cleaning scenarios are far from being well  evaluated~\cite{xiao2020benchmark, ahmadi2006multi,jager2002dynamic, 7361161, jeon2016multiple}.

In this paper, we aim at addressing these problems of {multi-robot hybrid task allocation in uncertain cleaning systems from both the perspectives of modeling and data. 
We explicitly model \emph{task uncertainty} and a set of practical constraints, such as the \emph{task-order constraint} which only allows allocating cleaning tasks to the robot cluster in specific orders. We then formulate this uncertain multi-robot hybrid-task allocation problem through robust mixed-integer linear programming} to minimize overall cleaning time.
Moreover, we investigate and discuss three types of uncertainty set, i.e., box, ellipsoidal and convex hull uncertainty set {to gain further insight into the role of the uncertainty set in modeling.}

{To compensate for the lack of public benchmark for cleaning task allocation, we establish} a benchmark, including a dataset and several baseline methods.
Firstly, we construct a dataset {for evaluating} the performance of different cleaning task allocation algorithms. This dataset contains $100$ {pairs of} images of manually labeled floor plans and simulated 3D environments for Robot Operating System (ROS). {The floor plans are selected from public dataset R3D~\cite{zeng2019deep}, which coverage a large various situations.}
Secondly, we conduct a thorough evaluation on the collected dataset. {A variety of solvers are evaluated to provide comprehensive baseline results,} including traditional optimization approaches including the simulated annealing algorithm, genetic algorithm, and particle swarm optimization, a deep reinforcement learning-based solver, and a commercial solver (Gurobi).
The benchmark data and baseline methods will be open-sourced upon the acceptance of the paper to further facilitate the related research via the link at \href{https://github.com/iamwangyabin/Multi-robot-Cleaning-Task-Allocation}{https://github.com/iamwangyabin/Multi-robot-Cleaning-Task-Allocation}.

The main contributions of this work can be summarized as:
\begin{itemize}
\item {We study the problem of uncertain multi-robot hybrid-task allocation for cleaning tasks and formulate it as a robust mixed integer linear programming model. To the best of our knowledge, this is the first study to consider and model the task uncertainty for cleaning task allocation.}

\item We construct a {dataset}, including manually labeled floor plans and 3D models from real floor plan images. The dataset covers various environments for practical applications. {To our best knowledge, it is the first dataset specifically designed for hybrid cleaning task allocation.}

\item We introduce {five baseline methods} for cleaning task allocation model and provide baseline results via comprehensive evaluation.
\end{itemize}

\section{Related Works}\label{sec2}

\subsection{Cleaning Robot Task Allocation}
According to different application scenarios, the multi-robots task allocation methods can be briefly split into two categories: the low-level planning and the high-level scheduling.
 \XP{A typical low-level allocation planning strategy is  the coverage path planning (CCP)~\cite{cabreira2019survey, almadhoun2019survey, lakshmanan2020complete}, which} tries to plan a feasible cleaning path for every robot to sweep all accessible area in a given region.
The formulation of the coverage path planning has been well studied, and \XP{become} the basic technique for cleaning robots.
Some researches, such as EA~\cite{le2020evolutionary}, ACO~\cite{han2020ant} and Network Flow~\cite{janchiv2013time}, formulate such low-level strategy as the well-known traveling salesman problem (TSP) or vehicle routing problem (VRP)~\cite{vidal2020concise, eshtehadi2020solving, guan2021kohonen}, where 
the cleaner has to visit all reachable regions with the minimum cost.
Recently, some other studies use reinforcement learning~\cite{lakshmanan2020complete, apuroop2021reinforcement, zhao2022task} to plan complete coverage paths for cleaning robots. 
Similar applications can be found for patrol and surveillance missions which continuously sweeps the area to detect interest~\cite{elmaliach2009multi, kolling2008multi}
\XP{However, as there is no standard and public benchmark,} these works were tested on specific environments and lack of the generalization and robustness for practical applications.
\XP{Moreover}, since such formulation is NP-hard, the computational time required to solve the problem increases drastically when the dimension of the problem grows. 

For large public environments, a more appropriate strategy \XP{for large-scale multi-robot task allocation is high-level scheduling}~\cite{ahmadi2006multi,jager2002dynamic, 7361161, jeon2016multiple, alitappeh2022multi}.
This strategy first partitions the whole workspace into several cleaning zones and assigns robots to these zones individually to minimize the total cost (time, distance, energy, etc.). Then, robots perform low-level coverage path planning individually in their assigned cleaning regions. This hierarchical optimization strategy \XP{is superior in achieving high-quality solution with reasonable computational costs, as the NP-hard path planning path is only performed in a small-scale space.}
Some common task allocation algorithms can be \XP{also} adopted to this scene, such as market-based methods~\cite{otte2020auctions}, optimization-based methods~\cite{yao2020online, banks2020multi, bai2021distributed} and behaviour-based methods~\cite{jin2019dynamic}.
\XP{The problem setting of large-scale multi-robot task allocation is relevant} to VRP problem~\cite{vidal2020concise, eshtehadi2020solving} or JSP problem~\cite{attiya2020job} \XP{as well}.
However, 
\XP{these approaches usually ignore}  the heterogeneous service ability of robots and the clean order constraints, which are, \XP{however,} critical factors for practical cleaning tasks. \XP{Instead, we take adequate account of these particularities and the proposed method can be directly} used for various types of cleaning robots task allocation problems, such as single-robot with single-cleaning tasks, and multi-robot single-cleaning tasks. 
 
Some strategies try to allocate specific coverage paths for robot cluster, which has a large computational complexity and difficult to deploy in real-world applications.
We only assign the cleaning zones to each robot in the cluster and let robots themselves to plan actual coverage path in the their assigned zones.
Our formulation makes the application of the task allocation more flexible and significantly reduce the computational complexity.

{There are only a few specific studies about cleaning robot task allocation.} Ahmadi and Stone~\cite{ahmadi2006multi} study the multi-cleaning robots task allocation in the large public environments, and point out that cleaning a large public space is a continuous area sweeping task.
In this work, they partition the space among robots, and multiple robots are assigned to their partitions. \XP{The study~\cite{7361161, jeon2016multiple} further} presents a mathematical formulation of cleaning a large public space with multiple robots, along with a procedural solution based on task reallocation.
\XP{These previous works only consider  \emph{single-task} allocation and are far away from practical application.}
At the same time, only few public datasets~\cite{xiao2020benchmark, inceoglu2018continuous} are provided for multi-robot task allocation problem, not to mention cleaning robot task allocation problem.

\subsection{Robust Optimization}
In general, robust optimization is \XP{to address the optimization problems against the worst instance with uncertain data.}
The robust optimization was first introduced to solve the linear optimization problem with uncertainty by Soystert~\cite{soyster1973convex} in 1973. 
MuIvey et al.~\cite{mulvey1995robust} proposed a general scenario-based robust optimization formulation that its solution is ``close" to optimal for all input scenarios.
Robust optimization assumes that the uncertain parameters belong to a given bounded uncertainty set, and several uncertainty sets (i.e. pure box; pure ellipsoidal; pure polyhedral) are studied in previous works~\cite{li2012robust, mulvey1995robust, ordonez2010robust}.
In general, the uncertainty set formulates to what degree of the solutions can violate the constraints, and thus the selection of the uncertainty set is critical for robust optimization.
In this paper, we mainly studies the influence of three classical uncertainty sets (pure convex, box and ellipsoidal uncertainty sets) on cleaning task allocation problem.

The rapid development of robust optimization theory has extended to several real-world applications, such as least squares problems~\cite{ei1997robust}, structural truss design~\cite{ben1997robust} and routing problem~\cite{tajik2014robust, saeedvand2019robust}.
The robust optimization assumes that the uncertain data belong to a bounded uncertainty set, but the parameters of that set are not needed.
Ben-Tal et al.~\cite{ben2006extending} proposed a new robust optimization methodology that has a controlled deterioration in performance when the data is outside the uncertainty set.
Bertsimas and Sim~\cite{bertsimas2004price} propose a robust optimization model which can flexibly adjust the level of conservatism of the robust solutions in terms of probabilistic bounds of constraint violations.
In this paper, we use robust optimization for the multi-robot cleaning task allocation problem. 
We follow the robust optimization proposed by Ben-Tal et al.~\cite{ben2006extending} and Li et al.~\cite{li2012robust}, and extend to the proposed multi-robot cleaning task allocation problem.

\section{Problem Formulation and Description}
In this section, we first provide the problem definition and present the mathematical formulation of our problem.
Then we consider the uncertainties in the large public environments and introduce robust optimization for the multi-robot cleaning task allocation problem. 

\begin{figure*}[h]
\centering
\includegraphics[scale=0.33]{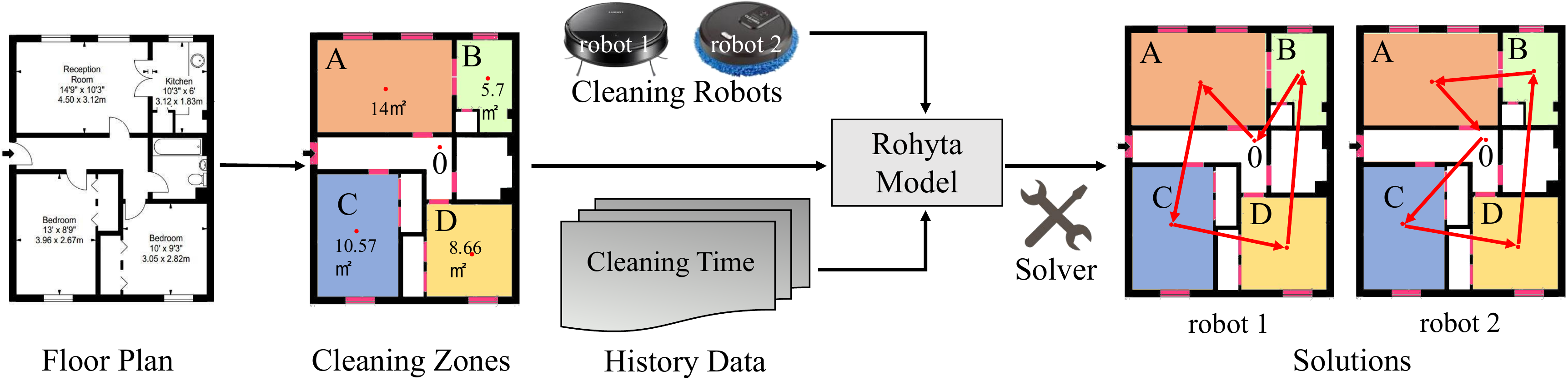}
\caption{Illustration of our proposed framework for uncertain multi-robot hybrid-task allocation problem. For a given floor plan, the area that needs cleaning is the cleaning zones, and labeled in different colors. Our proposed Rohyta model takes information about cleaning zones, cleaning robots and history data to produce robust solutions for task allocation. After getting the assignment, multiple heterogeneous cleaning robots will clean these zones orderly.}
\label{fig:framework}
\end{figure*}

\subsection{Problem Definition}
\label{sec:problem_def} 

Let $J$ be the set of $N$ cleaning tasks needs to be allocated. 
Given the set $R$ of $K$ heterogeneous cleaning robots, \XP{high-level multiple-cleaning-robot hybrid-task allocation algorithm assigns} robots \XP{in} $R$ to execute tasks \XP{within} $J$ orderly in such a way that the total cleaning time $C_{max}$ \XP{is minimized.}

Specifically, each task in $J$ has three attributes: \emph{location}, \emph{cleaning area}, and \emph{cleaning type}.
The \emph{location} of a task is where the task starts and ends, which is the centroid of the corresponding cleaning area.
The \emph{cleaning area} is the area of the task and can be obtained directly from floor plans.
The \emph{cleaning type} indicates the cleaning type of this task, such as vacuuming or mapping.
Cleaning robots have four attributes: \emph{moving speed}, \emph{robot ability}, \emph{maximum run-time}, and \emph{cleaning efficiency}.
The \emph{robot ability} indicates what types of cleaning work this robot can perform, because most robots are designed to do a specific single cleaning task.
It is corresponding to the aforementioned cleaning type of the cleaning task.
The \emph{moving speed} is the velocity of the robot travel between two places, which is an important factor used to calculate the time of robot switching between different cleaning tasks. 
The \emph{cleaning efficiency} means the area of a robot can clean per unit time, which is also an important parameter needs to consider.
The \emph{maximum run-time} means the maximum number of hours a robot can work without charge.

\XP{Clearly the magnitude $N$ of the task set $J$ is determined by the cleaning zones and the types of operators required for cleaning the zone (\emph{e.g.}, vacuuming, mopping, and disinfecting). An example can be found by referring to the given floor plan shown in the leftmost part of Fig.~\ref{fig:framework}, where four cleaning zones are labeled in different colors with A, B, C, and D.  
\XP{Suppose that for} every cleaning zone, we have to vacuum and then mop. As a result, there are $8$ cleaning tasks in total, with one vacuuming and one mopping task for each cleaning zone.}
{Note that} in our settings, all robots start from the depot, orderly execute the planned cleaning tasks, and finally return to the depot. 
\XP{Without losing generality, we append an auxiliary task with an area of $0$ (\emph{Task $0$}) in $J$ to indicate the depot.}
All robots shall start cleaning from this task, and finally come back to finish this task.

\subsection{Robust Multi-Robot Hybrid Task allocation Model}
\label{sec:robost_model}

In this section, we introduce the proposed RObust multi-robot HYbrid Task Allocation model~(Rohyta) for  Multiple-cleaning-Robot Hybrid-Task Allocation (MRHTA), as illustrated in Figure.~\ref{fig:framework}. 
Given an environment with several cleaning zones, we first obtain the attributes of cleaning tasks according to the floor plan. Note that the cleaning zones are predetermined. We then assign cleaning robots to clean these zones multiple times and record the cleaning time as historical data, which is used for robust optimization. Using this information, we apply the proposed ROHYTA model and optimization solvers to obtain the final solutions.

\begin{table}[]
\centering
\caption{Summary of variables}
\label{tab:notation}
\begin{tabular}{p{1cm} p{9cm}}
\toprule
$N$ & The total number of cleaning tasks.  \\
$K$ & The total number of cleaning robots. \\
$J$ & The set of $N$ cleaning tasks. Each task $j\in J$ is indexed numerically according to it's cleaning zone and cleaning task type.\\
$R$ & The set of $\XP{K}$ cleaning robots. Each robot $r \in R$ is indexed numerically.\\
\midrule
\multicolumn{2}{l}{\textbf{\XP{Pre-defined} Problem Parameters}}\\
$\mathbf{P}$ & The $N \times N$ \XP{binary} \emph{task precedence matrix}. $P_{i,j}= 1$ if task $i \in J$ should be done before $j \in J$. Otherwise  $P_{i,j}= 0$.\\
$\mathbf{B}$ & The $N \times K$ \XP{binary} \emph{robot ability matrix}. $B_{j, r}=1$ if robot $r \in R$ is able to complete task $j \in J$. Otherwise  $B_{j, r}=0$.\\
$\mathbf{D}$ & \XP{The $N \times K$} \emph{cleaning time consumption matrix}. $D_{j,r}$ is the cleaning time for robot $r\in R$ to complete cleaning job $j\in J$ and $D_{j,r} \in \mathcal{R}$.\\
$\mathbf{T}$ & The $N \times N \times K$ \emph{travel time \XP{array}}. Travel time needed for robot $r\in R$ between cleaning jobs $i$ and $j$, for $(i,j)\in J$ and $T_{i,j, r} \in \mathcal{R}$.\\
$\mathbf{L}$ & The maximum running time vector with a length of $K$, which indicates the maximum running time for each robot, and $L_{r} \in \mathcal{R}$. \\
$\lambda$ & A \XP{large-enough} constant number.\\
\hline
\multicolumn{2}{l}{\textbf{Variables \XP{to solve}}}\\
$\mathbf{X}$ & A $N \times N \times K$ \XP{binary \emph{edge assignment} array}. $X_{i,j,r}$ equals to $1$ if an edge $(i,j)$ is travelled by robot $r$, and $0$ otherwise.\\
$\mathbf{Y}$ & A $N \times K$ \XP{binary \emph{task assignment}} matrix. $Y_{j,r}$ equals to $1$ if cleaning task $j \in J$ is assigned to cleaning robot $r\in R$, and $0$ otherwise.\\
$C_{max}$ & The makespan of the task allocation schedule.\\
$U_{j}$ & The start time of the task $j \in J$.\\
\bottomrule 
\end{tabular}
\end{table}

Table.~\ref{tab:notation} summaries the variables used in Rohyta model.
The \emph{task order constraint matrix} $\mathbf{P}$ pre-defines the precedence and the order constraints of tasks. $P_{i,j}=1 \ \text{for} \ i \ \text{and} \ j \ \in J$ means task $i$ should be completed before task $j$. If there is no constraint between tasks $i$ and $j$, then $P_{i,j}=0$. 
To ensure efficient cleaning in a given zone that requires both vacuuming and mopping, the vacuuming task must be prioritized over the mopping task. This means that the robots should first vacuum the floor before mopping it. 
Additionally, since each robot in the heterogeneous cluster can perform different cleaning tasks, we need to impose constraints on the robots.
The \emph{robot ability matrix} $\mathbf{B}$ indicates what types of cleaning task a robot can perform.
If robot $r \in R$ can perform task $j \in J$, then $B_{j, r}=1$. Otherwise $B_{j, r}=\XP{0}$.
The \emph{cleaning time matrix} $\mathbf{D}$ indicates the time required for the robot to complete a cleaning task.
The matrix $\mathbf{D}$ is calculated according to the cleaning efficiency of robots and the cleaning area of tasks.
For tasks that the robot can't undertake, the cleaning time is set to $0$, because this item won't affect the final solution.
The \emph{travel time array} $\mathbf{T}$ is calculated by the shortest route between two \XP{tasks} and robot's travel speed.
The shortest route between two task's locations can be obtained by the shortest path algorithms, such as Dijkstra and a-star algorithm~\cite{hart1968formal}.
The maximum run time vector $\mathbf{l}$ is determined by robots' maximum running time, which is an attribute of robots.

Multi-robot cleaning task allocation problem tries to assign robots in the cluster according to their respective abilities and task requirements.
All robots start from depot, complete assigned tasks in turn, and return to depot.
The makespan $C_{max}$ of the whole cleaning work, which is also the total time spent on all cleaning tasks, is the time it takes from the first robot leaves the depot to the last robot returns. 
And the objective of the Rohyta model is to minimize this.
$X_{i,j,r}$ and $Y_{j,r}$ are \XP{binary} variables  to be solved.
$X_{i,j,r}$ equals to $1$ if the robot $r$ travels from task $i$ to task $j$, and $0$ otherwise.
$Y_{j,r}$ equals to $1$ if cleaning task $j \in J$ is assigned to cleaning robot $r\in R$, and $0$ otherwise.
$U_{j}$ is the start time of the cleaning task $j$, which is an auxiliary variable.

Given all variables above, the objective function and constraints of the Rohyta model are listed as follows.

\begin{align} 
\label{model:deterministic}
&\XP{\min} \quad  C_{max},  \\
\intertext{s.t.}
&C_{max} \geq U_i+D_{i,r}+T_{i,0,r} X_{i,0,r} , \forall i \in J \backslash \{0\}, r \in R \\
&Y_{i,r}=0 , \forall i \in J \backslash \{0\}, r \in R , B_{i,r}=0 \\
&\sum_{r \in R} Y_{0, r}=|R| , \forall i \in J \backslash \{0\}, \\
&\sum_{r \in R} Y_{i, r}=1 , \forall i \in J \backslash \{0\}, \\
&\sum_{i \in J \backslash \{0\}} X_{i, 0, r}=1 , \forall r \in R,
\\
&\sum_{i \in J} X_{i, j, r}=Y_{j, r} , \forall j \in J \backslash \{0\}, r \in R, \\
&\sum_{j \in J} X_{i, j, r}=Y_{i, r} , \forall i \in J \backslash \{0\}, r \in R, \\
&U_{i}+D_{i, r}+T_{i, j, r} \leq U_{j}+ \lambda \cdot (1-X_{i, j, r}), \forall r \in R, \forall (i,j) \in J, B_{j,k}\geq 0,  \\
&U_{j} \geq U_{i}+(D_{i, a}+T_{i, j, a}) \cdot (Y_{i, a}+Y_{j, b}-1), \forall (i,j)\in J,(a,b)\in R,i\neq j,\notag \\
&\quad\quad\quad\quad\quad\quad P_{i,j}=1,B_{i,a}=1,B_{j,b}=1, 
\\
&\sum_{i \in J} Y_{i, r} D_{i,r} < L_r , \forall r \in R.
\end{align}

The objective function is defined in $(1)$ and $(2)$, which ensure to minimize the overall completion time $C_{max}$.
Constraint $(3)$ states that a robot can serve a cleaning task only it has the ability to execute it. 
Constraint $(4)$ ensures that all the cleaning robots in depot should be allocated.
Constraint $(5)$ ensures that all cleaning tasks are aligned, and each job is served once.
Constraint $(6)$ make sure all robots return to the original depot at the end.
Constraints $(7)-(8)$ are the classical subtours elimination constraints, which is similar to traditional VRP problem.
Constraint $(9)$ state that if two cleaning tasks are assigned to the same robot, their execution times should not overlap.
Constraint $(10)$ makes sure that the start time of descendant cleaning should wait until the predecessor is completed.
Constraint $(11)$ makes sure that any robots can never work more than their maximum run time. 
In an ideal scenario, the cleaning environment is precisely known and the allocation algorithm will consistently produce the same solutions given the same initial state. 
However, in real-world scenarios, cleaners often face uncertain situations, particularly in large public environments. 
These uncertainties can cause a significant delay in cleaning time compared to the ideal value. 
Ignoring these uncertainties can lead to poor robot assignments and increased wait times.
Fortunately, historical data can be used to model such uncertainties and generate more robust solutions in practice.
We use robust optimization~\cite{soyster1973convex, ordonez2010robust} to model the uncertainty in the cleaning environment.
We assume the time consumption of the cleaning task belongs to a bounded uncertainty set $\Gamma$, as the following equation:
\begin{equation}
\label{eq:uncertain} 
\widehat D=\left\{\widehat D_{i,r} \mid \bar D_{i,r} +\sum_{s=1}^{S} \gamma_{s} \tilde D_{i,r}^{s}, \gamma \in \Gamma \right\}.
\end{equation}
Specially, the uncertain cleaning time $\widehat D_{i,r}$ is constructed by deviations $\tilde D_{i,r}^{s}$ around the ideal value $\bar D_{i,r}$. 
The deviations $\tilde D_{i,r}^{s} \in R^n$ ($s=1,2,...,S$) is a vector of history scenarios, which is the history records of deviation from the ideal consumption time $ \bar D_{i,r}$.
The ideal cleaning time is a theoretical value calculated from the cleaning area and robots' cleaning efficiency.
And the uncertain set $\widehat D$ is the bounded linear combination of scenario vectors around $\bar D_{i,r}$.
Different uncertainty set $\widehat D$ can generated by using different $\Gamma$ sets as Eq.~\ref{eq:gamma}.
\begin{equation}
\label{eq:gamma}
\begin{array}{c}
\Gamma_{1}=\left\{\gamma \in \mathcal{R}^{s} \mid \gamma \geq 0, \sum_{s=1}^{s} \gamma_{s} \leq 1\right\}, \\
\Gamma_{2}=\left\{\gamma \in \mathcal{R}^{s} \mid\|\gamma\|_{\infty} \leq 1\right\}, \\
\Gamma_{3}=\left\{\gamma \in \mathcal{R}^{s} \mid \gamma^{T} Q \gamma \leq \Omega^{2}\right\},
\end{array}
\end{equation}
where $\Gamma_{1}$, $\Gamma_{2}$ and $\Gamma_{3}$ refer to a Box Set, Convex Hull Set, and Ellipsoidal Set, respectively. The ellipsoidal set is defined using a positive definite matrix $Q$~\cite{sungur2008robust}, for example, $Q = I$. 
Following the robust optimization method~\cite{ordonez2010robust}, we can get the derivation $(14)-(16)$ for three uncertainty sets.
\begin{align}
&\widehat D_{j,r}= \bar D_{j,r}+\max \left\{\max _{S} \tilde D_{i,r}^{s}, 0\right\}, & \text { if } \Gamma=\Gamma_{1}, \\
&\widehat D_{j,r}= \bar D_{j,r}+\sum _{S} D_{i,r}^{s}, & \text { if } \Gamma=\Gamma_{2}, \\
&\widehat D_{j,r}= \bar D_{j,r}+\sqrt{\tilde D_{j,r,\cdot}^{T} Q^{-1} \tilde D_{j,r,\cdot}}, & \text { if } \Gamma=\Gamma_{3}, 
\end{align}

where $\tilde D_{j,r, \cdot}=(\tilde D_{j,r}^{1}, \tilde D_{j,r}^{2},\tilde D_{j,r}^{3},..., \tilde D_{j,r}^{s})$.

In the aforementioned Rohyta model, constraints $(2)$, $(9)$, $(10)$ and $(11)$ all have uncertain cleaning time parameter $D_{i,r}$. 
We can replace $D$ with $\widehat D_{j,r}$ respectively to get the robust MILP model.
If we don't consider uncertainties, we can also use the ideal cleaning time $\bar D$ to replace $D$.

\subsection{Solver}
For the proposed robust mixed-integer linear programming model, we need mathematical optimization solvers to get the solutions.
We first use a commercial programming solver (Gurobi) as our baseline, which is one of the fastest and best solvers.
But for large environments, we find Gurobi can not find feasible solutions within given time (one hour for example), and often produce bad solutions.
Thus, we design three classic intelligent algorithms and a deep reinforcement learning based solver, which are detailed in the following section.

\section{Benchmark Details}
\label{sec:Benchmark}
In this section, we detail our proposed multi-robot hybrid tasks allocation (MRHTA) benchmark, including baseline solvers and data.

\subsection{Baseline Optimization Solvers}
We introduce three intelligent algorithms, Simulated Annealing Algorithm (SA), Genetic Algorithm (GA) and Particle Swarm Optimization Algorithm (PSO), and a reinforcement learning based solvers 
as baseline methods to solve our robust MILP model.
Details of the implementation are provided in the following section.
A commercial mathematical optimization solver, Gurobi, is also one of our baseline methods, but we omit the implementation in this section for we just follow the product instructions.

In our implement, the representation of a feasible solution is a vector.
The first part of the vector is the service order of the cleaning zones, and if we have several cleaning works (vacuuming, mopping, ...), we list service order of all works in sequence.
The second part of the vector is the workload of each cleaning robot, that is the number of cleaning zones robots need to perform
For example, given a environment of three cleaning zones ($1,2,3$) and each cleaning zone has two cleaning works, vacuuming and mopping.
The heterogeneous robust cluster has two types of robots (vacuuming and mopping work), with two robots for each. 
We indicate these robots as R1, R2, R3 and R4, of which "R1, R2" represent vacuuming robots and "R3, R4" represent mopping robots. 
A feasible solution vector "$1,2,3\|3,2,1\|2,1\|1,2$" means robot R1 vacuums $1$ $2$ zones orderly, R2 vacuums zone $3$, R3 mops zone $3$ and R4 mops zone $2$ and $1$ orderly.
The solution representation above is also used for other intelligent optimization algorithms (GA and PSO).

\begin{figure}[!h]
\centering 
\includegraphics[scale=0.28]{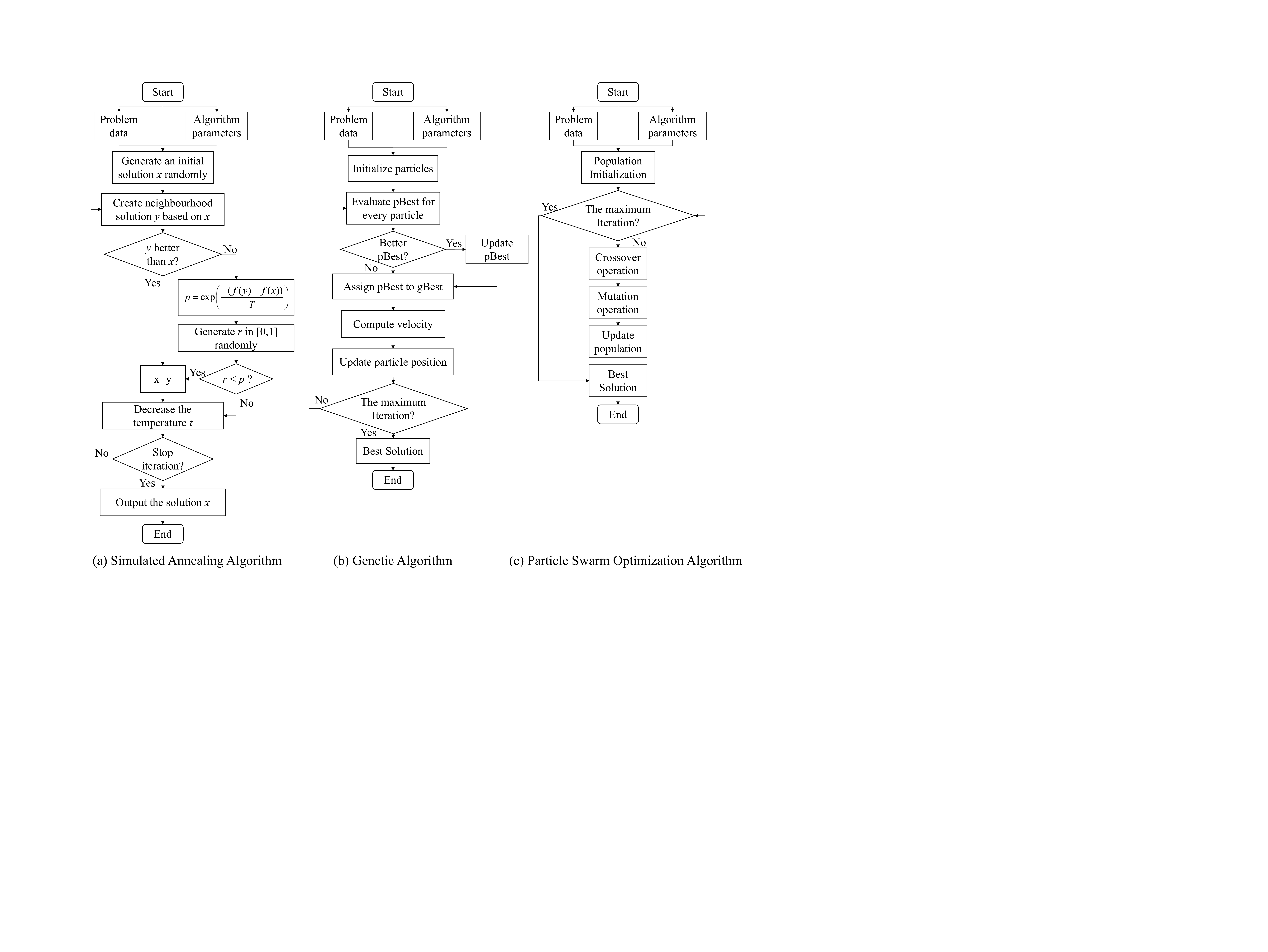}
\caption{Flow Chart of three Algorithms. (a) The flow chart of Simulated Annealing Algorithm. (b) The flow chart of Genetic Algorithm. (c) The flow chart of Particle Swarm Optimization Algorithm.}
\label{fig:algo_flow}
\end{figure}

\paragraph{Simulated Annealing Algorithm}
Simulated Annealing (SA)~\cite{kirkpatrick1983optimization} is an intelligence optimization algorithm for approximating the global optimum of a given system.
Simulated annealing is inspired by annealing in metallurgy where a material is heated to a high temperature and then slowly cool down to obtain strong crystalline structure.
The optimal solution in optimization corresponding to the minimum energy state in the annealing process.  
The search process in SA is similar as stochastic hill climbing but it gives the probability to get out of the local optima.
The implements of our simulated annealing to solve our model is shown in Fig.~\ref{fig:algo_flow} (a). 
Every iteration, a new feasible neighbor solution $y$ is generated randomly.
If the new neighbor solution is better than current solution $x$, this solution will be accepted.
Otherwise, it will go through metropolis criteria to judge whether to accept this sub-optimal solution or not.
The probability $p$ to accept a worse solution depends on the current temperature $T$ and energy degradation $-f(y)-f(x)$ of the objective value, as shown is Eq.~\ref{eq:prob_sa}.
A random value $r$ between $0$ and $1$ is generated every iteration. 
If $p>r$, the solution $y$ is accepted, otherwise it is rejected. 
\begin{equation}
\label{eq:prob_sa}
p=\exp \frac{-f(y)-f(x)}{T}.
\end{equation}

The operations to generate neighbor solutions are swap and reversion of the solution vector elements randomly.
In order to get feasible neighbor solutions, the algorithm repeats the random generation process until the solutions are feasible.
The SA intelligent algorithm uses four parameters, Iter, T0, Ts, and $\alpha$. 
$Iter$ denotes the number of iterations for which the search proceeds at a particular temperature, while $T0$ represents the initial temperature, and $Ts$ represents the final temperature, below which the SA procedure is stopped. 
Finally, $\alpha$ is the coefficient controlling the cooling schedule. 

\paragraph{Genetic Algorithm}
Genetic Algorithm (GA)~\cite{holland1992genetic} is one of the most basic evolutionary algorithms, which simulates Darwin's theory of biological evolution to get the optimum solution.
GA algorithm contains four main steps: Reproduction, Mutation, Crossover and Selection, and repeats these four steps for every iteration until convergence (the solution does not change).
The flow chart of GA is shown in Fig.~\ref{fig:algo_flow} (b).

GA algorithm starts from a set of individuals of a population, each individual is a feasible solution vector mentioned above.
We randomly generate these individuals until get feasible solutions.
The probability that an individual will be selected for reproduction is based on its makespan time $C_{max}$.
We use crossover and mutation operations to reproduce best solutions.
The crossover operation exchanges parts of two solutions with each other at a given crossover rate $G_c$.
The mutation operation randomly change some codes to others in a given solution at the mutation rate $G_m$.
The mutation and crossover operations are also repeated until we can get feasible new solutions.
Every iteration, new solutions are generated through the crossover and mutation of species in the population, and better solutions are selected for the next round of competition.

The GA intelligent algorithm contains four parameters, Iter, $P_s$, $G_c$, and $G_m$. 
$Iter$ denotes the number of iterations, while $P_s$ represents the population size, $G_c$ represents the crossover rate, which is the probability of crossover operator, and $G_m$ is the probability of mutation operator. 

\paragraph{Particle Swarm Optimization Algorithm}
Particle swarm optimization (PSO)~\cite{bansal2019particle} is an evolutionary computation optimization algorithm, which was inspired by the behaviour of flocks of birds and herds of animals. 
The basic idea of PSO is to find the optimal solution through collaboration and information sharing between individuals in a population.
PSO algorithm starts from a set of particles, with the number of $N_p$.
In PSO, each particles in the population only has two properties: speed $v^{t}_{i}, i \in N_p$ and position $p_{i}^{t}, i \in N_p$, with speed represents the speed of particle's movement and position represents the direction of movement.
Each particle individually searches for the local optimal solution (pBest) in its search space, which is the best solution so far by that particle.
And particles share the individual best position with the other particles in the whole swarm to get the current global optimal solution (gBest) for the whole swarm. 
And all particles in the swarm adjust their speeds (under the Maximum velocity $V_{max}$) and positions according to gBest and pBest.
Particles' velocities $V_{t}$ is updated as Eq.~\ref{eq:pso update v}
\begin{equation}
\label{eq:pso update v}
v^{t+1}_{i} = W \dot v^{t+1}_{i} + c_1 U_{1}^{t}(pBest^t_{i}-p_{i}^{t}) + c_2 U_{2}^{t}(gBest-p_{i}^{t}),
\end{equation}
where $W$ is the Inertia weight, $c_1$ is the cognitive constant, $c_2$ is the social constant, $U_1$ and $U_2$ are random numbers. 
And moving particles to their new positions as Eq.~\ref{eq:pso move p}.
\begin{equation}
\label{eq:pso move p}
p_{i}^{t+1} = p_{i}^{t} + v^{t+1}_{i},
\end{equation}
PSO is convergent if the velocity of the particles will decrease to zero or stay unchanged.

The flow chart of PSO is shown in Fig.~\ref{fig:algo_flow} (c).
For cross-border processing of particles, we simply set these particles to the border values.
Parameters of PSO algorithm are listed as follows: The number of particle $N_p$, The number of iteration $Iter$, and the Maximum velocity $V_{max}$.

\begin{figure*}[!h]
\centering 
\includegraphics[scale=0.4]{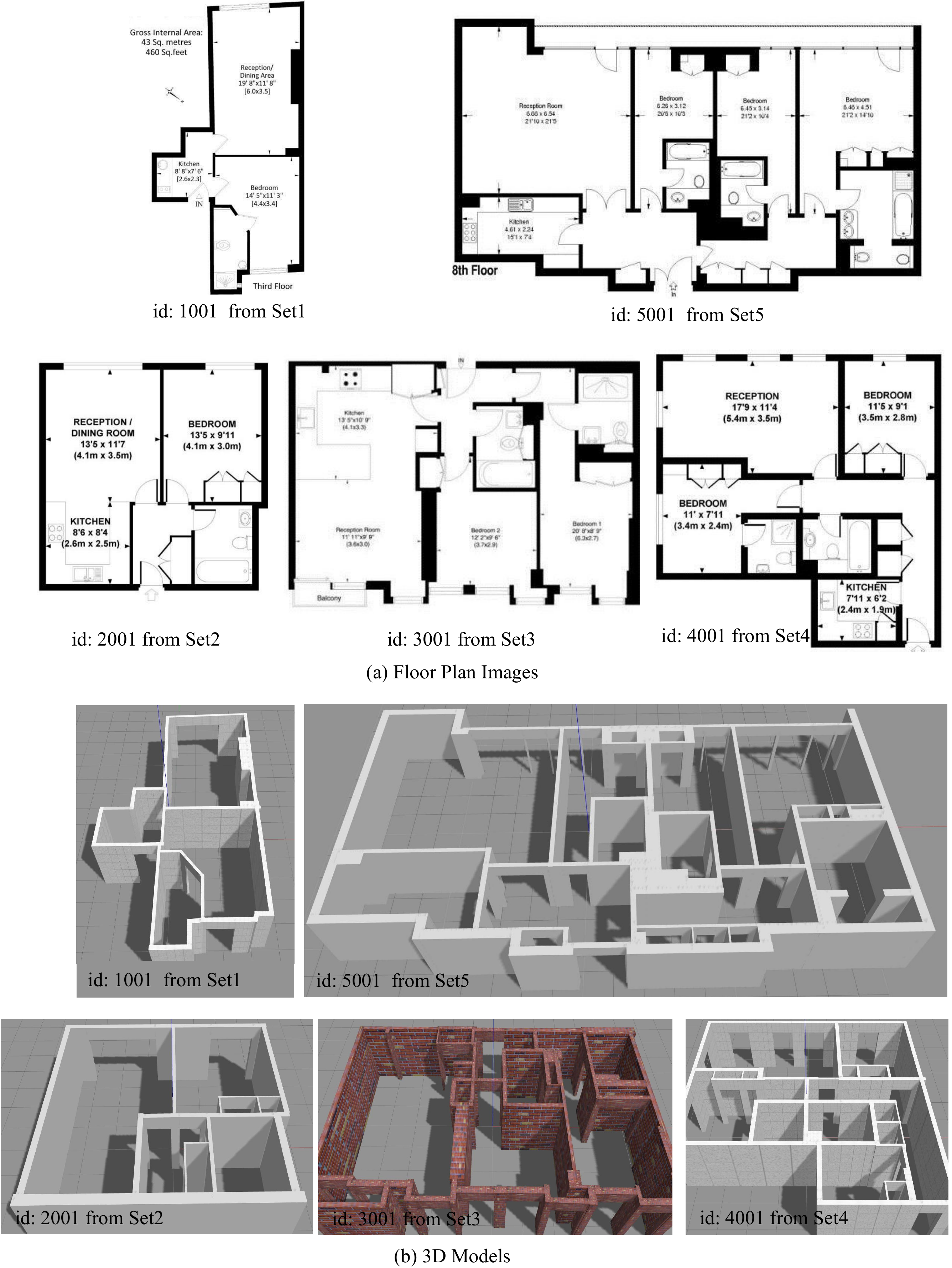}
\caption{Illustration of our dataset. (a) The floor plan images from~\cite{liu2015rent3d,zeng2019deep}. (b) The corresponding 3D models under the same scale. Material and texture of different rooms are visualized in picture. From set1 to set5, the area of the environment and the number of rooms increase in turn.}
\label{fig:samples}
\end{figure*}

\begin{figure*}[t]
\centering
\includegraphics[scale=0.4]{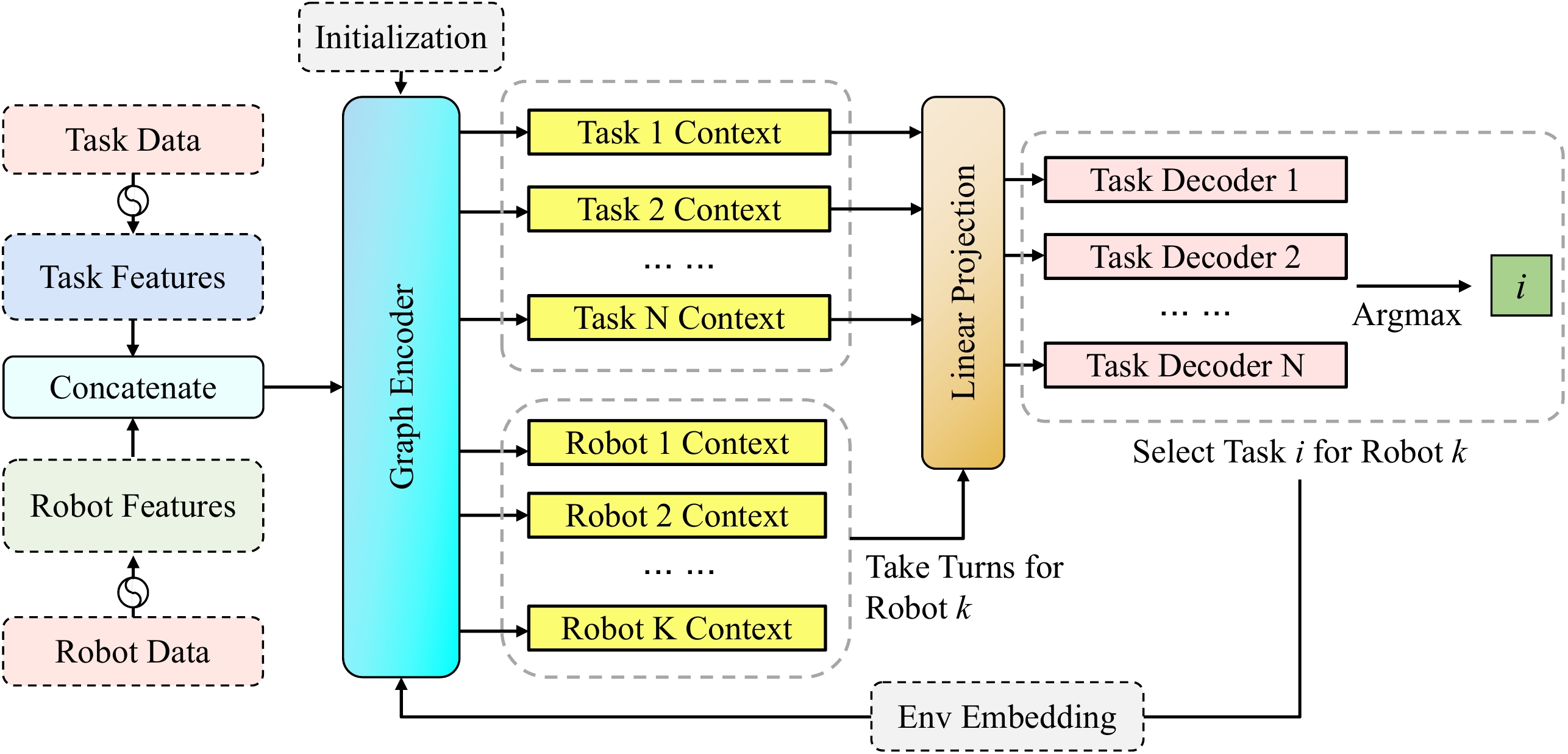}
\caption{Illustration of \XP{the proposed end-to-end reinforcement learning based solver.} The RL solver has two main parts. The left part is a set of embeddings that encode the problem data (Task and Robot related data) into high-dimension features. The Graph Encoder use the context-based attention mechanism~\cite{10.5555/3327546.3327651} to produce context embeddings. The second component is a RNN decoder that points to an input at every decoding step.   
Every decoding step, the decoder chooses a robot in turn and outputs a task for this robot.
}
\label{fig:network}
\end{figure*}

\paragraph{Reinforcement learning}
Given the success of deep neural networks in computer vision and natural language processing, recently there has been a trend to use reinforcement learning (RL) to tackle optimization problems.
Due to the properties of our task allocation problem, we choose to use an end-to-end reinforcement learning based solver~\cite{10.5555/3327546.3327651} as our baseline.
The architecture of our proposed RL solver is illustrated in Fig.~\ref{fig:network}.

Our network is a Transformer model, which use an encoder-decoder architecture.
The encoder takes Task Data, Robot Data and Environment Embedding as input. 
The Task Data contains main attributes of cleaning zones (region area and distance between two regions), which is a vector, and we use a linear function to embed the structure data into a high dimension vector($N \times 128 $ in our experiments).
The Robot Data contains contain robot's cleaning efficiency, travel speed and the service ability, and we use the same way to get the robot feature.
The job precedence constraint is implicitly declared through the calculation of reward function.

The other part of the policy model is the same as~\cite{10.5555/3327546.3327651}.
The whole process is similar to the human decision-making process.
The network takes the state from the environment (Env Embedding) and gathers all problem information to make a decision of current step.
Then previous decision would yield a new environment state, and the network use the updated information to make a decision of next step.
But the origin policy network is designed for VRP problem, which only has one vehicle.
In their implements, the network would produce all nodes sequentially until there is no node remains.
However, for our heterogeneous cleaning robot task allocation, it is unsuitable to produce tour sequentially.
We let the input of the decoder is an embedding of the current region property and the robot's embedding.
Every step, the network choose a robot in turn and yield a task for this robot.
The reward function is $C_max$, which is mentioned in~\ref{sec:robost_model}.

\subsection{Dataset Details}
Currently, many different algorithms~\cite{7361161, jeon2016multiple} have been proposed for multi-robot task allocation problem, but one of the biggest challenges for the research community is the absence of large public datasets for evaluation, especially for cleaning robots.
To evaluate the performance of the task-allocation algorithms for multiple cleaning robots in real-world application, we propose a large public dataset.
For task allocation algorithm, robustness and optimal are the most important measurement indexes.
Different algorithms require different types of simulation environment for evaluation, for example coverage path planning generally tested on 2D grid maps, our cleaning zone based allocation only needs major information of cleaning zones, and some works require simulated 3D environments.
In order to meet various needs, we use real high-quality real floor plans to generate our synthetic data.
The floor plan images are from public dataset R3D~\cite{zeng2019deep}, which has 214 images with pixel-wise labeled floor plan recognition.
These images are original from \cite{liu2015rent3d}, and has manually labeled, like walls, doors, bedrooms, living room, etc.
We further select $100$ images and manually calibrate these images to actual size (pixel to meter).
What's more, we also manually construct virtual 3D environments for simulation.
Considering the significant differences among these floor plans, we divided the entire dataset into five subsets according to floor plan's attributes, mainly based on the number of rooms and the area of the house.
The details of the dataset is listed in table.~\ref{tab:dataset}. 
Fig.~\ref{fig:samples} illustrates five instances from our dataset, including the original floor plans and the 3D models. 

\begin{table}[h]
\caption{Details of the dataset.}
\label{tab:dataset}
\centering \small
\renewcommand{\arraystretch}{1.5}
\begin{tabular}{p{0.25\textwidth}|c|c|c|c|c}
\hline
Subset    & Set1     & Set2 & Set3 & Set4 & Set5 \\
\hline
{\#}floor plans    & 23       & 25   & 23   & 23   & 6    \\
\hline
{\#}cleaning zones & 3$\sim$6 & 7$\sim$9  & 10$\sim$12   & 12$\sim$18   & $18 \sim 35$   \\
\hline
Average area ($m^2$) & 38.2     & 62.3   & 73.0   & 85.9   & 148.0  \\
\hline
Cleaning types	       & 2        & 2    & 2    & 2    & 2 \\
\hline
\end{tabular}
\end{table}

We intend to provide an easy interface to facilitate researchers verifying their algorithms.
The experiment environment, dataset and implementation of our five solvers are open-source, and we also provide example codes of other algorithms, such as coverage path planning.
The 3D model can be used to test cleaning robots using ROS and Gazebo.

\section{Experiments}
In this section, we first present our experiment settings in the Sec.~\ref{sec:Experiment settings}, including our performance measurements and algorithm settings.
Then we show the results of our task allocation algorithm on one example instance in Sec.~\ref{sec:Analysis of Deterministic Solution}.
In Sec.~\ref{sec:Performance Comparison}, we show all results of our dataset, including the objective values and computing time of five solvers.
Finally, we analyze the results of our robust solutions in Sec.~\ref{sec:Analysis of Robust Solution}.

\subsection{Experiment settings}
\label{sec:Experiment settings}
In our experiments, we need first partition the whole clean area into several cleaning zones.
Most of floor plans in our dataset are partitioned based on room function, such as kitchen and bedroom.
We also manually split some large rooms into smaller cleaning zones, which can get more reasonable assignment. 
It should be noted that the goal of the cleaning task allocation is to complete the whole room cleaning as quickly as possible.
The partition strategy of the cleaning zones can be changed to suit the algorithms, and our implements can be seen as a baseline method.
Because the clean area of Set1 is too small to use all robots, we only use robot 1 and robot 3 in the table.~\ref{tab:robots}. 
For the rest data, we use all robots in the table.~\ref{tab:robots}.
All cleaning \XP{zones} need two cleaning tasks: vacuuming and mopping, and vacuuming task should be completed before mopping.
In order to get the moving time between two cleaning zones, we use a-star algorithm~\cite{hart1968formal} to find the shortest path on the grid map, and use the robot's moving speed to calculate the moving time.
The major properties of cleaning robots in our implements are listed in Table.~\ref{tab:robots}.

\begin{table}[]
\caption{Summary of major properties of cleaning robots.}
\label{tab:robots}
\centering \small
\renewcommand{\arraystretch}{1.5}
\begin{tabular}{p{0.25\textwidth}|c|c|c|c}
\hline
Robot Name           & Robot1    & Robot2       & Robot3  & Robot4       \\
\hline
Robot Ability        & Vacuuming & Vacuuming    & Mopping & Mopping      \\
\hline
Covering Area($m^2$) & 150       & 200          & 100     & 100          \\
\hline
Runtime($h$)           & 2.5       & 3.0$\sim$3.5 & 2       & 2.5$\sim$3.5 \\
\hline
Battery Capacity($mA$) & 3200      & 5200         & 2150    & 2300         \\
\hline
Travel Speed($m/s$)    & 0.2       & 0.2          & 0.2     & 0.2          \\
\hline
Cleaning Speed($m^2/s$) & 0.016     & 0.023        & 0.04    & 0.07        \\
\hline
\end{tabular}
\end{table}

Tree intelligent optimization algorithms, a deep reinforcement learning based solver (RL Solver) and a commercial mathematical optimization solver (Gurobi) are used to solve our problem.
Generally, intelligent algorithms and RL Solver can not obtain global optimum solution and has certain degree of randomness, but they can give relative good results within the fixed time.
Gurobi can give the global optimum solution if it has enough time, and we set 10 minutes as the upper bound.
After several experiments, we decide to fix the parameters as follows.
The parameter values for Genetic Algorithms (GA): $Iter = 3,000$, $P_s = 200$, $G_c =  0.9$ and $G_m = 0.08$.
The parameter values for Simulated Annealing (SA): $T0$ = $500$, Ts = $1$, $\alpha = 0.997$, $Lk = 300$, $Iter = 3,000$.
PSO parameters are listed as follows: Number of particle $N_p$ is $2,000$, Number of maximum iteration is $1000$ and the Maximum velocity $V_{max}$ is $2$. Inertia weight $W$ is $0.5$, the cognitive constant $c_1$ and the social constant $c_2$ are $1$. 
Particles' initial speed $v^{t}_{i}$ and position $p_{i}^{t}$ initialized randomly.

We train the RL Solver using the same parameters as \cite{10.5555/3327546.3327651}, except we use much more epochs($50,000$).
That is probably because our problem is more complicated than VRP.
We generate random data for training RL Solver (including the number and area of regions and distances between them).
After training, we test the RL solver directly on the proposed dataset.
All the methods are coded in Python and tested with the varieties of five subsets(Set1, Set2, Set3, Set4 and Set5 in table.~\ref{tab:dataset}). 
And experiments are carried out with a runtime limit of one hour on a workstation with two Intel Xeon E5-2690v3, two GTX 2080 super GPU and 64G RAM running Windows 10 system. 

To test the performance of different solvers, we first conduct experiments on the deterministic environment, which uses the ideal cleaning time ($\bar D_{j,r}$) in the Rohyta model.
And we name it as the deterministic solution.
For robust solutions, we have to change the ideal cleaning time ($\bar D_{j,r}$) to the uncertain cleaning time($\widehat D_{j,r}$).
Since we don't have real history cleaning time, we generate them randomly and use different deviations to simulate the real world condition.
We randomly generate a total of $10$ scenarios within the given percent deviation for the three uncertainty sets.
The percent values of cleaning time deviation are as follows: $5\%$, $10\%$, $15\%$.
We use the Simulated Annealing (SA) solver for our robust model, because it's comprehensive performance is the best.
To measure the performance of robust solutions, we use the ratio $r_{ro}$ to quantify the relative extra cost of the robust with respect to the cost of the deterministic, and the $r_{ro}$ is given as:
\begin{equation}
r_{ro}=\frac{C_{max}^{r} - C_{max}^{d}}{C_{max}^{d}}.
\end{equation}
$C_{max}^{r}$ is the objective function value of the robust solution, $C_{max}^{d}$ is the objective function value of the deterministic solution.
This ratio gives information on how much extra cost we will incur if we want to implement the robust to protect against the worst case realization of the uncertainty, instead of implementing the deterministic. 

\subsection{Analysis of Deterministic Solution}
\label{sec:Analysis of Deterministic Solution}
\begin{figure}[htbp]
    \centering
    \subfigure[Optimal]{
        \includegraphics[width=5.6cm]{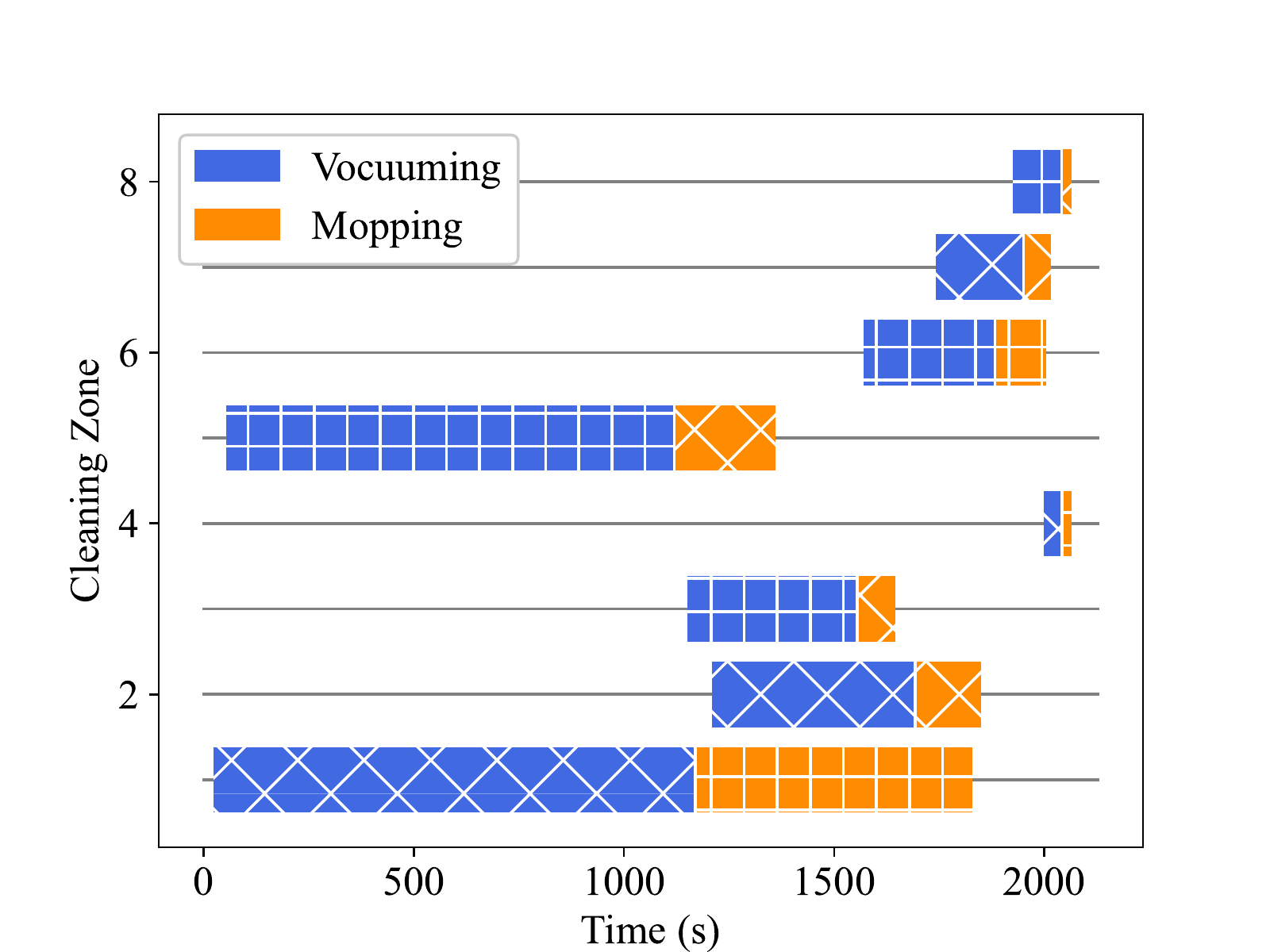}
    }
    \subfigure[Gurobi]{
	\includegraphics[width=5.6cm]{Fig/gantt/gurobi.pdf}
    }
    \quad   
    \subfigure[SA]{
    	\includegraphics[width=5.6cm]{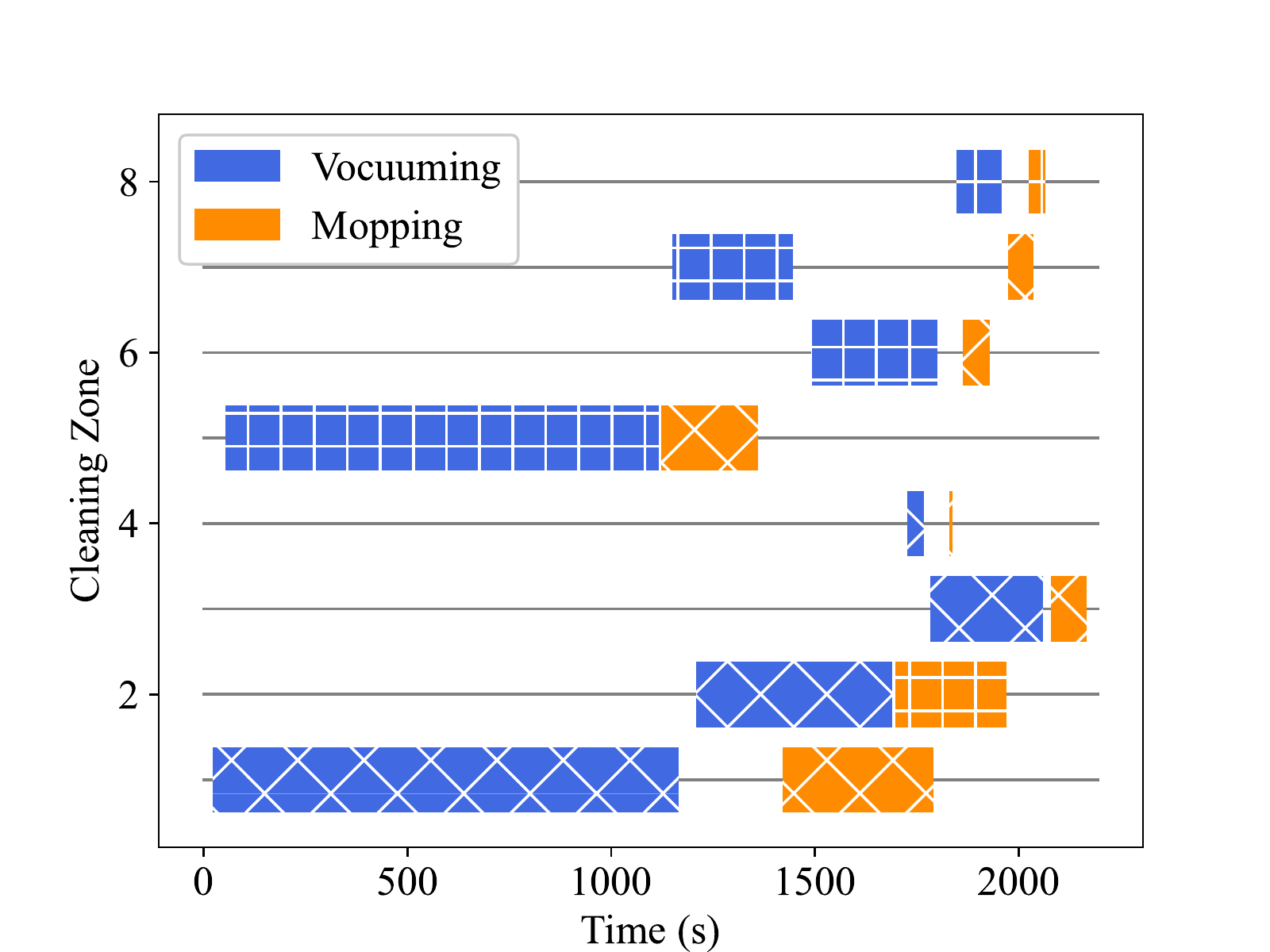}
    }
    \subfigure[GA]{
	\includegraphics[width=5.6cm]{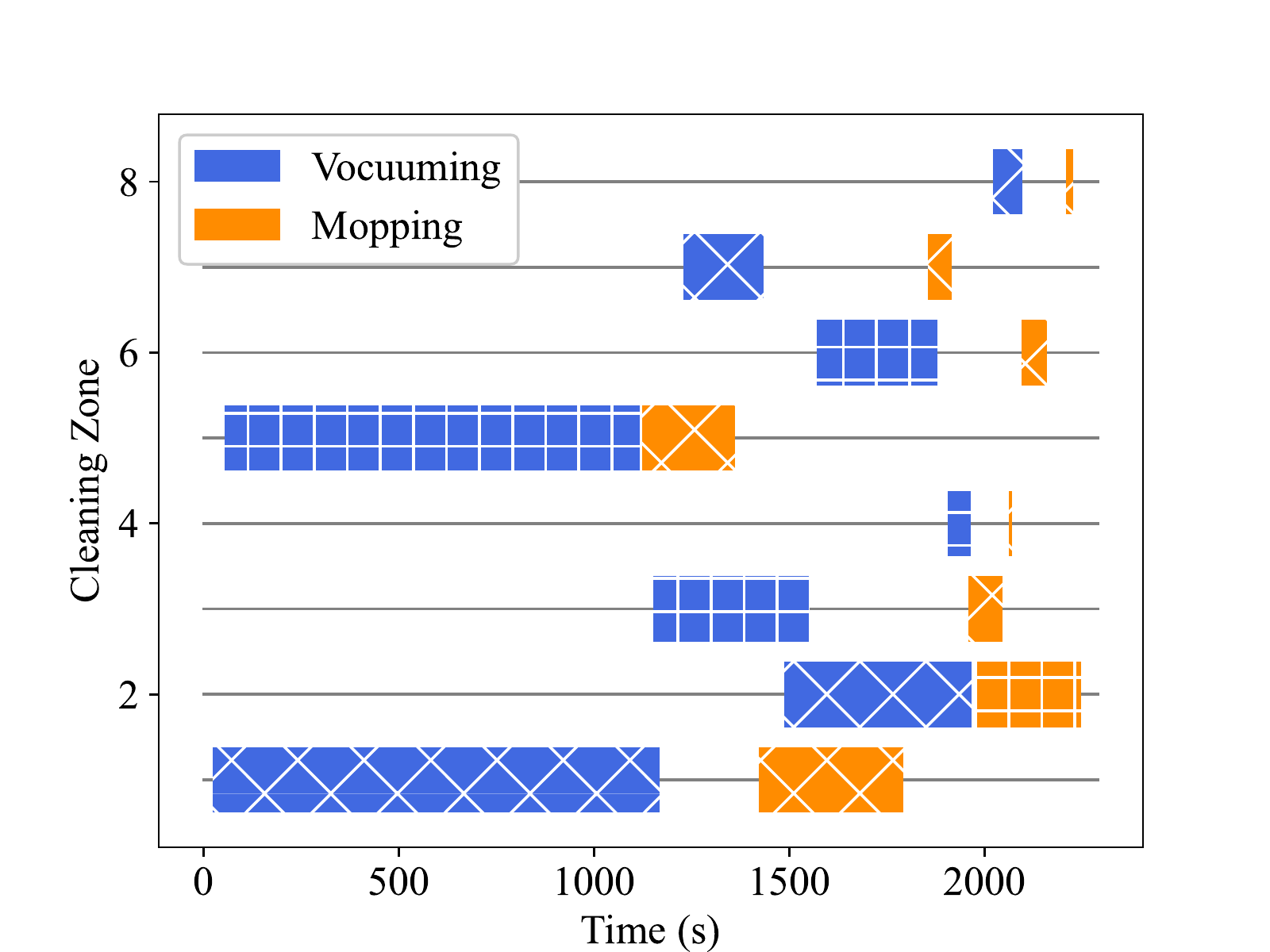}
    }
    \quad   
    \subfigure[PSO]{
    	\includegraphics[width=5.6cm]{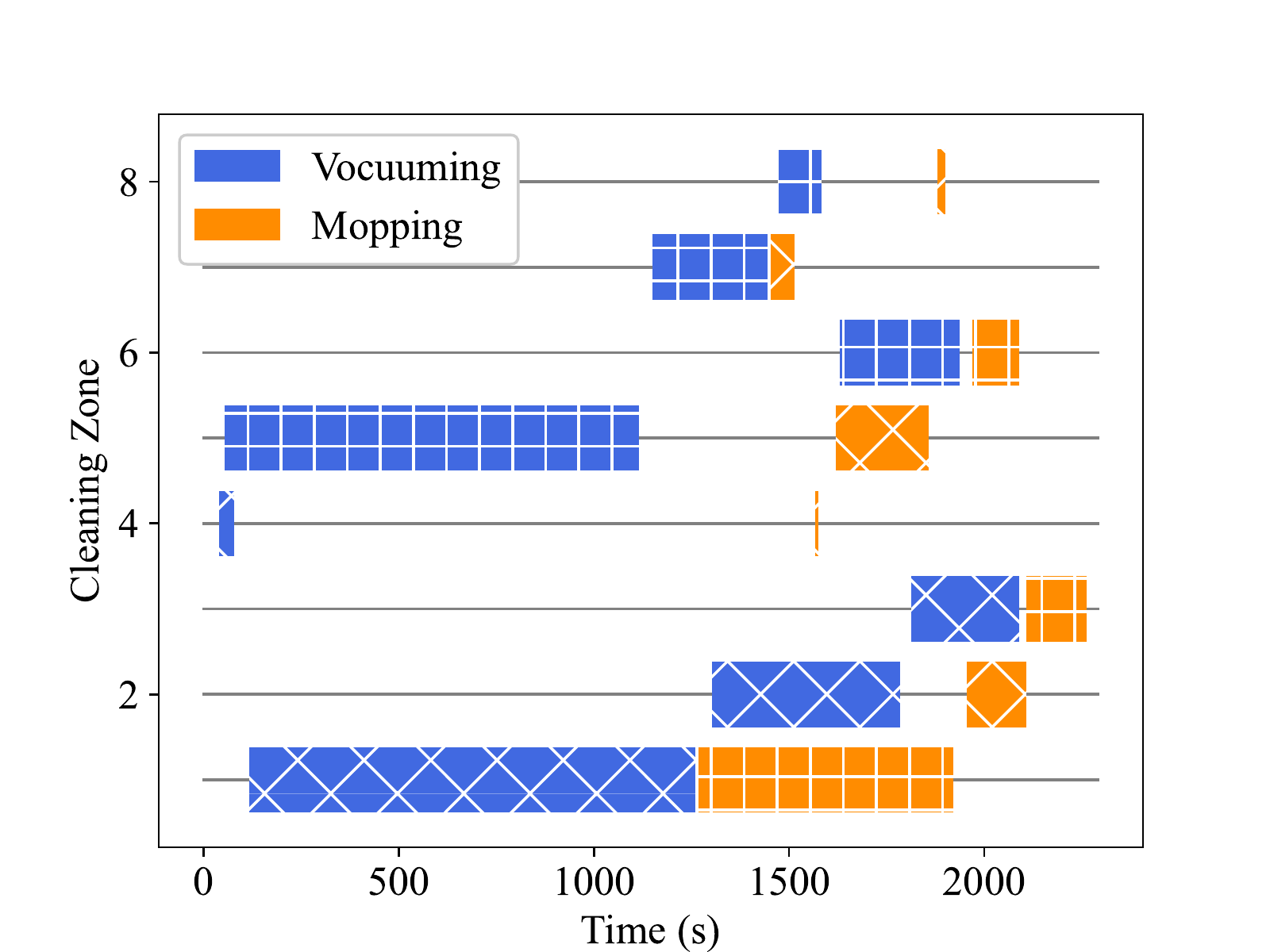}
    }
    \subfigure[RL]{
	\includegraphics[width=5.6cm]{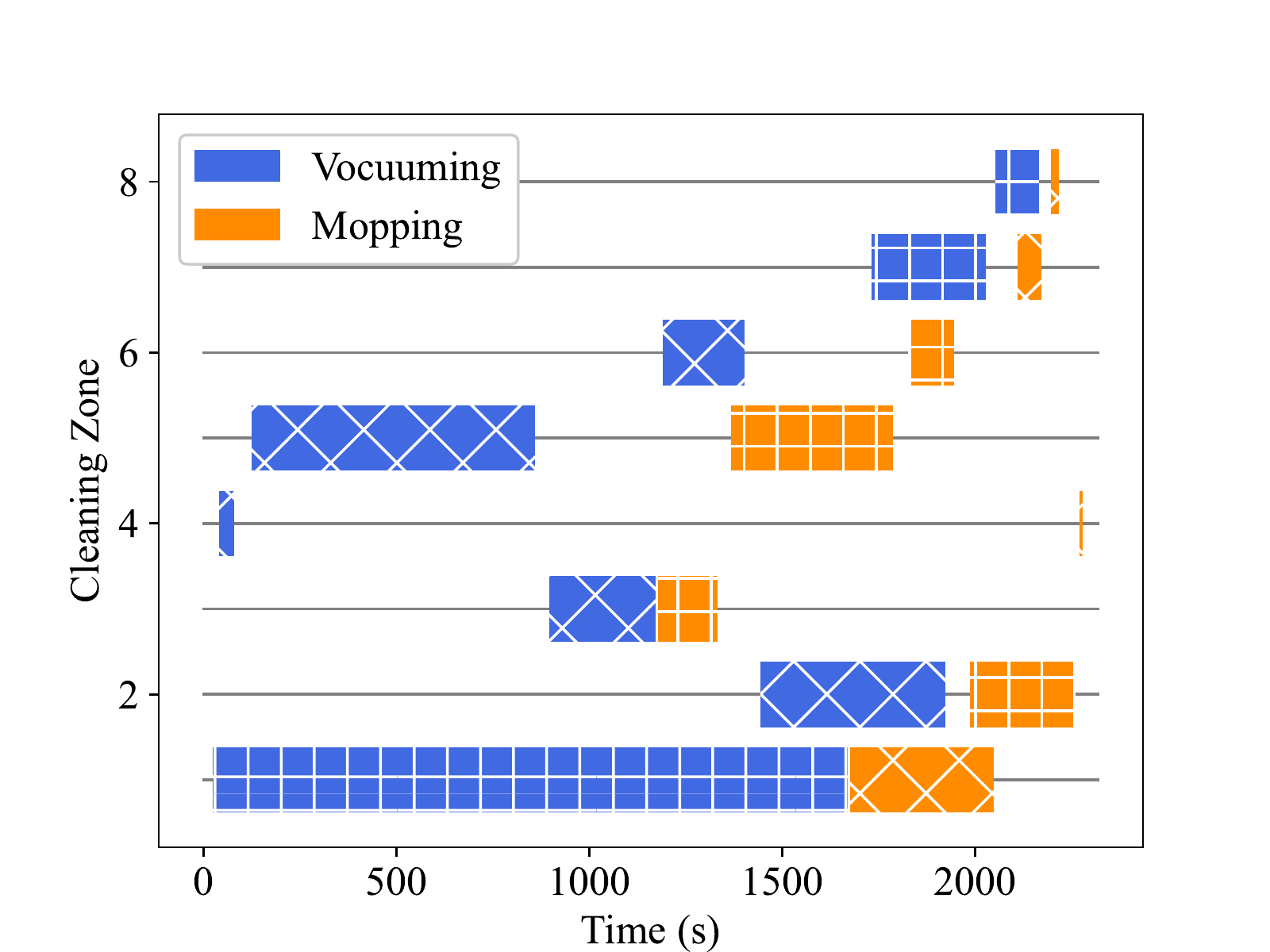}
    }
    \caption{Gantt charts of five solvers for a given instance (id:2010). The x-axis is the timeline. The y-axis stands for cleaning zones. The progress, duration, the start and the end time of each region for every robot are shown. The different patterns of the same color in the figure represent different robots belonging to the same cleaning ability.
}
    \label{fig:instance}
\end{figure}

We first use one instance (id.2010 from set2) from our dataset to show the result of our task allocation model.
The results of our task allocation including the order of the cleaning jobs, the start time and end time of each cleaning jobs, and travel paths for cleaning robots to switch between cleaning zones. 
The visualization of our deterministic solution of five solvers (GA, SA, PSO, Gruobi, RL) are shown in Fig.~\ref{fig:instance}.

For this instance (id. 2010), the objective value of the Gurobi solver is $2128s$, the SA is $2195s$, the GA is $2290s$, the RL Solver is $2320s$, and the PSO is $2292s$.
The optimum solution is obtained by Gurobi solver, and other solvers are easily produce suboptimal solutions.
As shown in Fig.~\ref{fig:instance}, Mopping robot can only clean a room when it has been vacuumed, and sometimes it inevitably has to wait.
The total number of cleaning zone is $8$, and the gantt chart shows the start and end time of each cleaning work (blue line is vacuuming and orange line is mopping) in all cleaning zones.
We can see that only the previous task (Vacuuming) of a zone is completed, then robots can mop this zone.
Optimal solutions can keep robots as busy as possible and thus minimize the total cleaning time.

Since the optimal value needs to minimize the wait time, any uncertainty of predecessor tasks may lead to the prolongation of the waiting time of successor tasks.
Thus if we take the uncertainty of cleaning time into consideration, the optimal solutions could become sub-optimal.

\subsection{Performance Comparison of Deterministic Model: Benchmark Instances} 
\label{sec:Performance Comparison} 
\begin{figure*}[ht]
\centering
\includegraphics[scale=0.35]{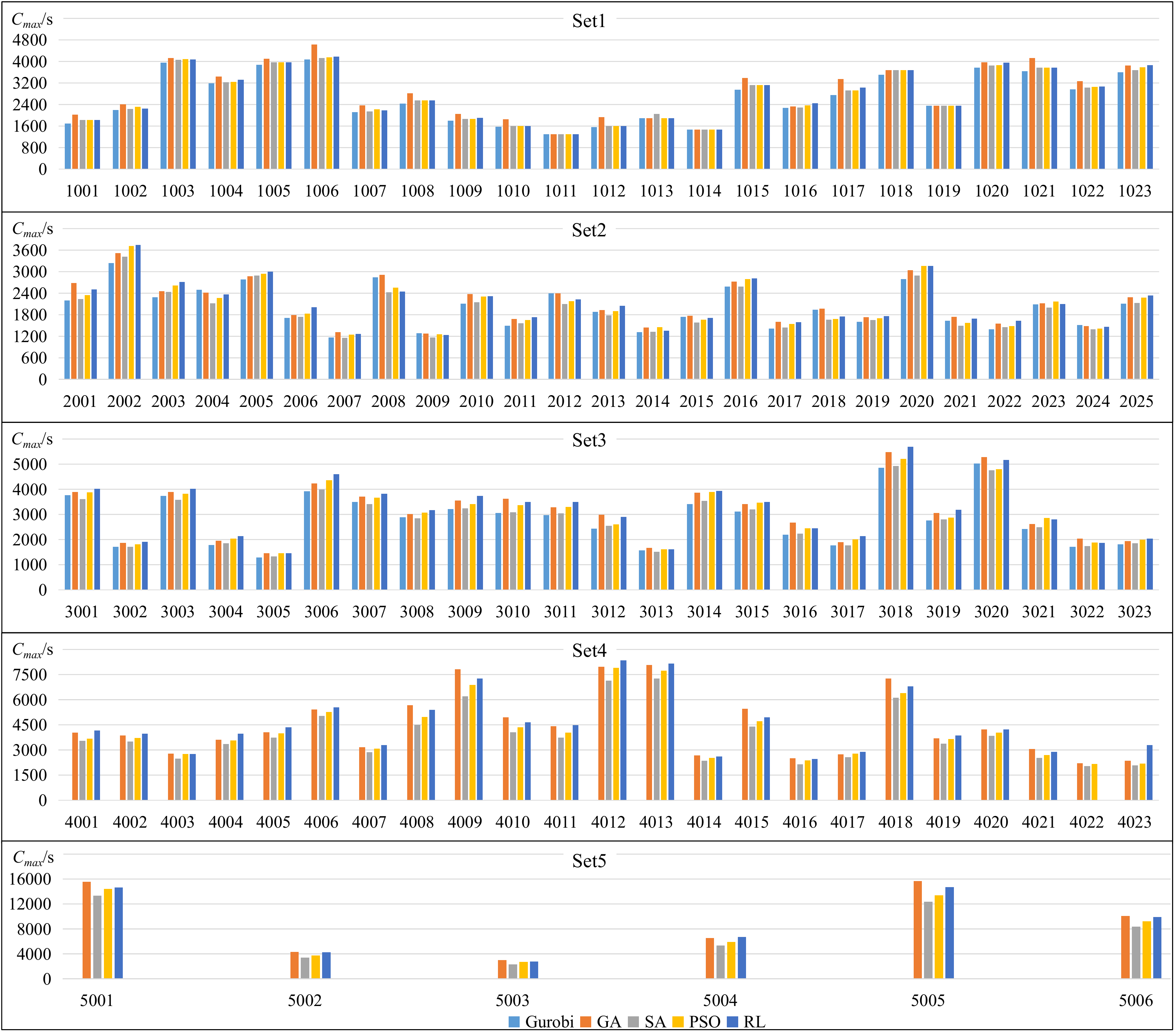}
\caption{The objective values of each instances in our dataset. The x-axis is instance ids. The y-axis is $C_{max}$.}
\label{fig:acc}
\end{figure*}

\begin{figure*}[ht]
\centering
\includegraphics[scale=0.35]{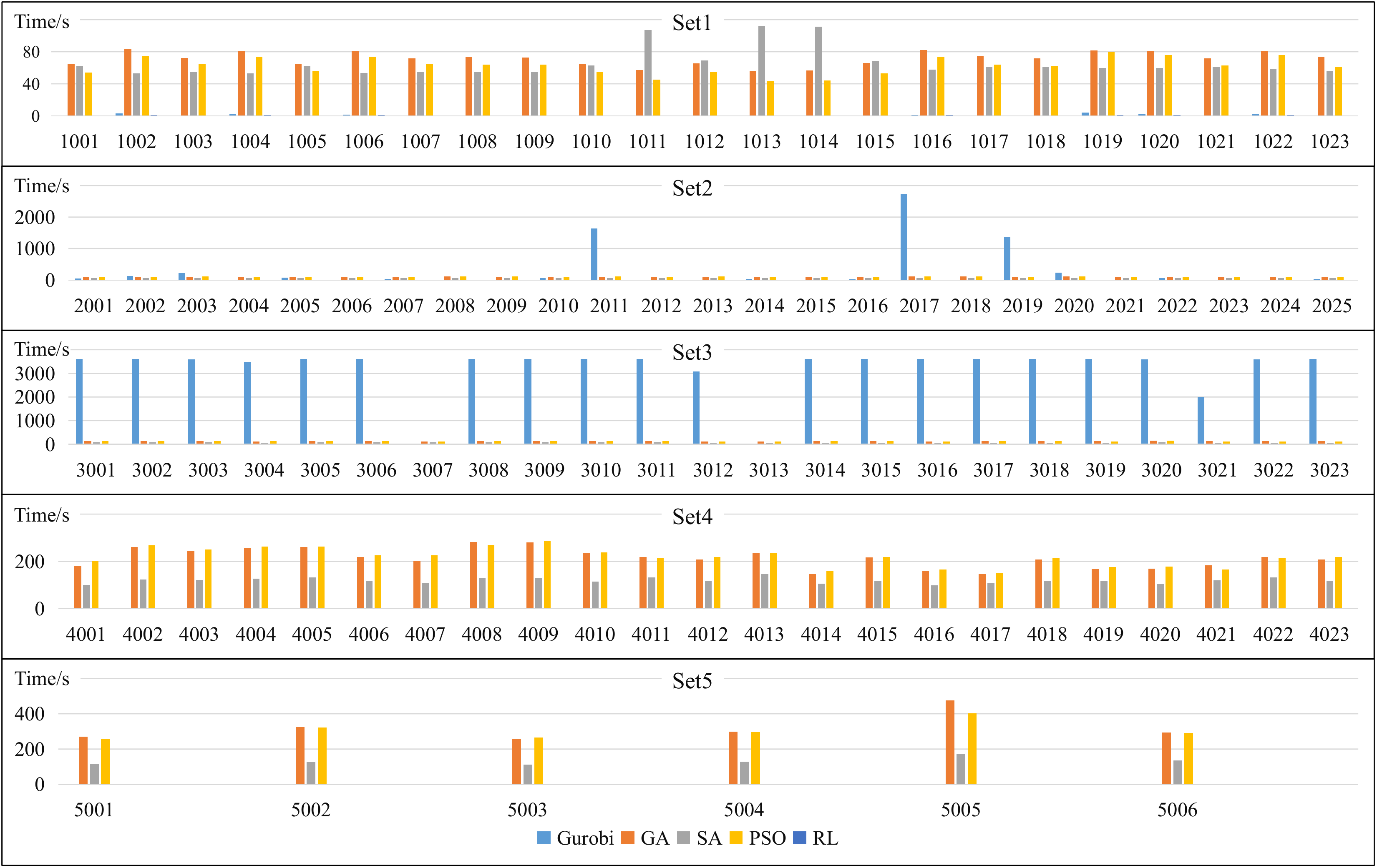}
\caption{The time consumption for solvers. The x-axis is instance ids. The y-axis is time spent to get solutions.}
\label{fig:time}
\end{figure*}

In the large public environment, garbage is constantly generated, and the demand for cleaning is also constantly generated, which requires the task allocation algorithms produce proper solutions in the given time.
So good solvers should have high precision and speed in all conditions.
We further calculate the deterministic solutions of all instances in our dataset, and evaluate the performance four solvers (Gurobi, SA, GA, PSO and RL).
The dataset contains various indoor rooms, which can fully measure the performance of our algorithms.
The evaluation metrics include objective value $C_{max}$ and time complexity (time-consuming).  
The test results of objective value ($C_{max}$) are shown in Fig.~\ref{fig:acc}, and the results of time-consuming are shown in Fig.~\ref{fig:time}.

\begin{table}[ht]
\centering
\begin{tabular}{crrrr}
\hline
\multirow{2}{*}{Uncertainty Set}   & \multicolumn{1}{c}{\multirow{2}{*}{Datasets}} & \multicolumn{3}{c}{Deviation Degree}  \\ \cline{3-5} 
&\multicolumn{1}{c}{}                          & 5\%     & 10\%    & 15\%    \\ \hline
\multirow{5}{*}{Box Set}         & Set1     & 14.39\% & 26.18\% & 38.05\% \\
& Set2 & 8.68\%  & 19.24\% & 30.15\% \\
& Set3 & 11.60\% & 23.21\% & 33.91\% \\
& Set4 & 10.22\% & 20.47\% & 30.04\% \\
& Set5 & 10.68\% & 21.19\% & 31.56\% \\ \hline
\multirow{5}{*}{Convex Hull Set} & Set1     & 8.62\%  & 14.62\% & 20.70\% \\
& Set2 & 3.34\%  & 8.85\%  & 14.03\% \\
& Set3 & 5.94\%  & 12.45\% & 17.98\% \\
& Set4 & 5.15\%  & 10.73\% & 15.51\% \\
& Set5 & 5.74\%  & 10.81\% & 16.87\% \\ \hline
\multirow{5}{*}{Ellipsoidal Set} & Set1     & 6.13\%  & 9.31\%  & 12.45\% \\
& Set2 & 0.19\%  & 3.25\%  & 6.23\%  \\
& Set3 & 4.01\%  & 7.05\%  & 8.23\%  \\
& Set4 & 2.25\%  & 5.42\%  & 7.34\%  \\
& Set5 & 3.74\%  & 5.66\%  & 9.62\%  \\ \hline
\end{tabular}
\caption{The cost ratio $r_{ro}$ varies across different uncertainty sets for robust solutions.}
\label{tab:robustresults}
\end{table}

For objective function, SA performs best in large environments (set4, set5), but in small environment the Gurobi solver can produce optimal solutions.
As a whole, the time consumption of three intelligent solvers is similar.
However, SA can always give better results than other solvers.
For large environments, Gurobi solver needs even days to get the optimum solution, otherwise, it gives sub-optimal results.
Even in some small instances, Gurobi solver needs relatively much more time than others.
Surprisingly, deep reinforcement learning based solver is much more bad than all above solvers, especially we use hundred hours (about one week) to train this network.
One of the reasons probably is the RL solver~\cite{10.5555/3327546.3327651} is originally designed to solver VRP problem, and our multi-robot task allocation problem is too hard for it.
What's more, consider the training time, RL solver is currently unsuitable to deploy in practice.
Considering both objective value and time complexity, SA is recommended to used in real-world applications.

\subsection{Analysis of Robust Solution} 
\label{sec:Analysis of Robust Solution}
If the environment of cleaning robots is static and deterministic, the deterministic formulation of task allocation is enough for practical applications.
However, the rage of uncertainties is unusually wide in large public environments, such as moving people and bad conditions of robots.
As shown in Sec.~\ref{sec:Analysis of Deterministic Solution}, the time consumption of any tasks is important, for it influence all next series of tasks.

We study the effect of uncertainty sets on the robust results, and choose the most suitable for our task.
The scenario vectors we use here is randomly generated based on the degree of deviation.
To imitate real world situations, we test three degrees ($5\%$, $10\%$, $15\%$) of deviation.
We use the SA solver to investigate three uncertainty set (convex, box and ellipsoidal uncertainty sets), and \XP{report the results} in Table.~\ref{tab:robustresults}.
We can see that all three can get robust solutions, but different uncertainty sets have different conservative degrees.

In the three uncertainty sets, the convex set is the least conservative and the box set is the most conservative under the same degree of deviation.
In a simple scenario, the numerical value fluctuates greatly, possibly due to the large denominator.
For cleaning task, overly conservative tactics is inefficient and totally impractical to deploy.
The robust solutions using Ellipsoidal Set can always maintained at a relatively low extra time consumption.
Thus the ellipsoidal uncertainty set is proper to use in practice.

\section{Conclusions}
In this work, we aim at addressing \XP{multi-robot hybrid task allocation in uncertain cleaning environment} from both the perspectives of modeling and data. 
We explicitly model the task uncertainty and a set of practical constraints using robust mixed-integer linear programming. 
To compensate for the lack of public benchmark for cleaning task allocation, we establish a benchmark, including a dataset and several baseline methods.
The results show that our problem formulation and optimization solvers can be used in practical applications.
The proposed benchmark can be also used to evaluate other cleaning task allocation algorithms.

However, the cleaning zones in our work are pre-defined, which is \XP{still sub-optimal for many situations}. In our future work, we will consider more factors to our model and use more allocation strategies.

\section{Data Availability Statement}
The data supporting the findings of this study have been made openly available through our provided dataset. Access to both the codes and data is publicly available at the following location: \href{https://github.com/iamwangyabin/Multi-robot-Cleaning-Task-Allocation}{}.

\section{Acknowledgments}
This work is funded by National Key Research and Development Project of China under Grant No. 2019YFB1312000 and by National Natural Science Foundation of China under Grant No. 62076195.

\bibliography{sn-article}


\begin{thebibliography}{49}
\ifx \bisbn   \undefined \def \bisbn  #1{ISBN #1}\fi
\ifx \binits  \undefined \def \binits#1{#1}\fi
\ifx \bauthor  \undefined \def \bauthor#1{#1}\fi
\ifx \batitle  \undefined \def \batitle#1{#1}\fi
\ifx \bjtitle  \undefined \def \bjtitle#1{#1}\fi
\ifx \bvolume  \undefined \def \bvolume#1{\textbf{#1}}\fi
\ifx \byear  \undefined \def \byear#1{#1}\fi
\ifx \bissue  \undefined \def \bissue#1{#1}\fi
\ifx \bfpage  \undefined \def \bfpage#1{#1}\fi
\ifx \blpage  \undefined \def \blpage #1{#1}\fi
\ifx \burl  \undefined \def \burl#1{\textsf{#1}}\fi
\ifx \doiurl  \undefined \def \doiurl#1{\url{https://doi.org/#1}}\fi
\ifx \betal  \undefined \def \betal{\textit{et al.}}\fi
\ifx \binstitute  \undefined \def \binstitute#1{#1}\fi
\ifx \binstitutionaled  \undefined \def \binstitutionaled#1{#1}\fi
\ifx \bctitle  \undefined \def \bctitle#1{#1}\fi
\ifx \beditor  \undefined \def \beditor#1{#1}\fi
\ifx \bpublisher  \undefined \def \bpublisher#1{#1}\fi
\ifx \bbtitle  \undefined \def \bbtitle#1{#1}\fi
\ifx \bedition  \undefined \def \bedition#1{#1}\fi
\ifx \bseriesno  \undefined \def \bseriesno#1{#1}\fi
\ifx \blocation  \undefined \def \blocation#1{#1}\fi
\ifx \bsertitle  \undefined \def \bsertitle#1{#1}\fi
\ifx \bsnm \undefined \def \bsnm#1{#1}\fi
\ifx \bsuffix \undefined \def \bsuffix#1{#1}\fi
\ifx \bparticle \undefined \def \bparticle#1{#1}\fi
\ifx \barticle \undefined \def \barticle#1{#1}\fi
\bibcommenthead
\ifx \bconfdate \undefined \def \bconfdate #1{#1}\fi
\ifx \botherref \undefined \def \botherref #1{#1}\fi
\ifx \url \undefined \def \url#1{\textsf{#1}}\fi
\ifx \bchapter \undefined \def \bchapter#1{#1}\fi
\ifx \bbook \undefined \def \bbook#1{#1}\fi
\ifx \bcomment \undefined \def \bcomment#1{#1}\fi
\ifx \oauthor \undefined \def \oauthor#1{#1}\fi
\ifx \citeauthoryear \undefined \def \citeauthoryear#1{#1}\fi
\ifx \endbibitem  \undefined \def \endbibitem {}\fi
\ifx \bconflocation  \undefined \def \bconflocation#1{#1}\fi
\ifx \arxivurl  \undefined \def \arxivurl#1{\textsf{#1}}\fi
\csname PreBibitemsHook\endcsname

\bibitem{cabreira2019survey}
\begin{barticle}
\bauthor{\bsnm{Cabreira}, \binits{T.M.}},
\bauthor{\bsnm{Brisolara}, \binits{L.B.}},
\bauthor{\bsnm{Ferreira~Jr}, \binits{P.R.}}:
\batitle{Survey on coverage path planning with unmanned aerial vehicles}.
\bjtitle{Drones}
\bvolume{3}(\bissue{1}),
\bfpage{4}
(\byear{2019})
\end{barticle}
\endbibitem

\bibitem{almadhoun2019survey}
\begin{barticle}
\bauthor{\bsnm{Almadhoun}, \binits{R.}},
\bauthor{\bsnm{Taha}, \binits{T.}},
\bauthor{\bsnm{Seneviratne}, \binits{L.}},
\bauthor{\bsnm{Zweiri}, \binits{Y.}}:
\batitle{A survey on multi-robot coverage path planning for model
  reconstruction and mapping}.
\bjtitle{SN Applied Sciences}
\bvolume{1}(\bissue{8}),
\bfpage{1}--\blpage{24}
(\byear{2019})
\end{barticle}
\endbibitem

\bibitem{lakshmanan2020complete}
\begin{barticle}
\bauthor{\bsnm{Lakshmanan}, \binits{A.K.}},
\bauthor{\bsnm{Mohan}, \binits{R.E.}},
\bauthor{\bsnm{Ramalingam}, \binits{B.}},
\bauthor{\bsnm{Le}, \binits{A.V.}},
\bauthor{\bsnm{Veerajagadeshwar}, \binits{P.}},
\bauthor{\bsnm{Tiwari}, \binits{K.}},
\bauthor{\bsnm{Ilyas}, \binits{M.}}:
\batitle{Complete coverage path planning using reinforcement learning for
  tetromino based cleaning and maintenance robot}.
\bjtitle{Automation in Construction}
\bvolume{112},
\bfpage{103078}
(\byear{2020})
\end{barticle}
\endbibitem

\bibitem{10.5555/3463952.3463974}
\begin{bchapter}
\bauthor{\bsnm{Aziz}, \binits{H.}},
\bauthor{\bsnm{Chan}, \binits{H.}},
\bauthor{\bsnm{Cseh}, \binits{A.}},
\bauthor{\bsnm{Li}, \binits{B.}},
\bauthor{\bsnm{Ramezani}, \binits{F.}},
\bauthor{\bsnm{Wang}, \binits{C.}}:
\bctitle{Multi-robot task allocation-complexity and approximation}.
In: \bbtitle{Proceedings of the 20th International Conference on Autonomous
  Agents and MultiAgent Systems}.
\bsertitle{AAMAS '21},
pp. \bfpage{133}--\blpage{141}.
\bpublisher{International Foundation for Autonomous Agents and Multiagent
  Systems},
\blocation{Richland, SC}
(\byear{2021})
\end{bchapter}
\endbibitem

\bibitem{shang2020co}
\begin{barticle}
\bauthor{\bsnm{Shang}, \binits{Z.}},
\bauthor{\bsnm{Bradley}, \binits{J.}},
\bauthor{\bsnm{Shen}, \binits{Z.}}:
\batitle{A co-optimal coverage path planning method for aerial scanning of
  complex structures}.
\bjtitle{Expert Systems with Applications}
\bvolume{158},
\bfpage{113535}
(\byear{2020})
\end{barticle}
\endbibitem

\bibitem{huang2018multiple}
\begin{barticle}
\bauthor{\bsnm{Huang}, \binits{L.}},
\bauthor{\bsnm{Ding}, \binits{Y.}},
\bauthor{\bsnm{Zhou}, \binits{M.}},
\bauthor{\bsnm{Jin}, \binits{Y.}},
\bauthor{\bsnm{Hao}, \binits{K.}}:
\batitle{Multiple-solution optimization strategy for multirobot task
  allocation}.
\bjtitle{IEEE Transactions on Systems, Man, and Cybernetics: Systems}
\bvolume{50}(\bissue{11}),
\bfpage{4283}--\blpage{4294}
(\byear{2018})
\end{barticle}
\endbibitem

\bibitem{7361161}
\begin{bchapter}
\bauthor{\bsnm{{Jeon}}, \binits{S.}},
\bauthor{\bsnm{{Jang}}, \binits{M.}},
\bauthor{\bsnm{{Lee}}, \binits{D.}},
\bauthor{\bsnm{{Cho}}, \binits{Y.}},
\bauthor{\bsnm{{Lee}}, \binits{J.}}:
\bctitle{Multiple robots task allocation for cleaning a large public space}.
In: \bbtitle{2015 SAI Intelligent Systems Conference (IntelliSys)},
pp. \bfpage{315}--\blpage{319}
(\byear{2015}).
\doiurl{10.1109/IntelliSys.2015.7361161}
\end{bchapter}
\endbibitem

\bibitem{ahmadi2006multi}
\begin{bchapter}
\bauthor{\bsnm{Ahmadi}, \binits{M.}},
\bauthor{\bsnm{Stone}, \binits{P.}}:
\bctitle{A multi-robot system for continuous area sweeping tasks}.
In: \bbtitle{Proceedings 2006 IEEE International Conference on Robotics and
  Automation, 2006. ICRA 2006.},
pp. \bfpage{1724}--\blpage{1729}
(\byear{2006}).
\bcomment{IEEE}
\end{bchapter}
\endbibitem

\bibitem{jeon2016multiple}
\begin{bchapter}
\bauthor{\bsnm{Jeon}, \binits{S.}},
\bauthor{\bsnm{Jang}, \binits{M.}},
\bauthor{\bsnm{Lee}, \binits{D.}},
\bauthor{\bsnm{Cho}, \binits{Y.-J.}},
\bauthor{\bsnm{Kim}, \binits{J.}},
\bauthor{\bsnm{Lee}, \binits{J.}}:
\bctitle{Multiple robots task allocation for cleaning and delivery}.
In: \bbtitle{Intelligent Systems and Applications},
pp. \bfpage{195}--\blpage{214}.
\bpublisher{Springer}, \blocation{???}
(\byear{2016})
\end{bchapter}
\endbibitem

\bibitem{jiang2019group}
\begin{botherref}
\oauthor{\bsnm{Jiang}, \binits{J.}},
\oauthor{\bsnm{An}, \binits{B.}},
\oauthor{\bsnm{Jiang}, \binits{Y.}},
\oauthor{\bsnm{Zhang}, \binits{C.}},
\oauthor{\bsnm{Bu}, \binits{Z.}},
\oauthor{\bsnm{Cao}, \binits{J.}}:
Group-oriented task allocation for crowdsourcing in social networks.
IEEE Transactions on Systems, Man, and Cybernetics: Systems
(2019)
\end{botherref}
\endbibitem

\bibitem{wang2021hybrid}
\begin{botherref}
\oauthor{\bsnm{Wang}, \binits{Z.}},
\oauthor{\bsnm{Xu}, \binits{Z.}},
\oauthor{\bsnm{Liu}, \binits{B.}},
\oauthor{\bsnm{Zhang}, \binits{Y.}},
\oauthor{\bsnm{Yang}, \binits{Q.}}:
A hybrid cleaning scheduling framework for operations and maintenance of
  photovoltaic systems.
IEEE Transactions on Systems, Man, and Cybernetics: Systems
(2021)
\end{botherref}
\endbibitem

\bibitem{xiao2020benchmark}
\begin{botherref}
\oauthor{\bsnm{Xiao}, \binits{K.}},
\oauthor{\bsnm{Lu}, \binits{J.}},
\oauthor{\bsnm{Nie}, \binits{Y.}},
\oauthor{\bsnm{Ma}, \binits{L.}},
\oauthor{\bsnm{Wang}, \binits{X.}},
\oauthor{\bsnm{Wang}, \binits{G.}}:
A benchmark for multi-uav task assignment of an extended team orienteering
  problem.
arXiv preprint arXiv:2009.00363
(2020)
\end{botherref}
\endbibitem

\bibitem{jager2002dynamic}
\begin{bchapter}
\bauthor{\bsnm{Jager}, \binits{M.}},
\bauthor{\bsnm{Nebel}, \binits{B.}}:
\bctitle{Dynamic decentralized area partitioning for cooperating cleaning
  robots}.
In: \bbtitle{Proceedings 2002 IEEE International Conference on Robotics and
  Automation (Cat. No. 02CH37292)},
vol. \bseriesno{4},
pp. \bfpage{3577}--\blpage{3582}
(\byear{2002}).
\bcomment{IEEE}
\end{bchapter}
\endbibitem

\bibitem{zeng2019deep}
\begin{bchapter}
\bauthor{\bsnm{Zeng}, \binits{Z.}},
\bauthor{\bsnm{Li}, \binits{X.}},
\bauthor{\bsnm{Yu}, \binits{Y.K.}},
\bauthor{\bsnm{Fu}, \binits{C.-W.}}:
\bctitle{Deep floor plan recognition using a multi-task network with
  room-boundary-guided attention}.
In: \bbtitle{Proceedings of the IEEE/CVF International Conference on Computer
  Vision},
pp. \bfpage{9096}--\blpage{9104}
(\byear{2019})
\end{bchapter}
\endbibitem

\bibitem{le2020evolutionary}
\begin{barticle}
\bauthor{\bsnm{Le}, \binits{A.V.}},
\bauthor{\bsnm{Nhan}, \binits{N.H.K.}},
\bauthor{\bsnm{Mohan}, \binits{R.E.}}:
\batitle{Evolutionary algorithm-based complete coverage path planning for
  tetriamond tiling robots}.
\bjtitle{Sensors}
\bvolume{20}(\bissue{2}),
\bfpage{445}
(\byear{2020})
\end{barticle}
\endbibitem

\bibitem{han2020ant}
\begin{barticle}
\bauthor{\bsnm{Han}, \binits{G.}},
\bauthor{\bsnm{Zhou}, \binits{Z.}},
\bauthor{\bsnm{Zhang}, \binits{T.}},
\bauthor{\bsnm{Wang}, \binits{H.}},
\bauthor{\bsnm{Liu}, \binits{L.}},
\bauthor{\bsnm{Peng}, \binits{Y.}},
\bauthor{\bsnm{Guizani}, \binits{M.}}:
\batitle{Ant-colony-based complete-coverage path-planning algorithm for
  underwater gliders in ocean areas with thermoclines}.
\bjtitle{IEEE Transactions on Vehicular Technology}
\bvolume{69}(\bissue{8}),
\bfpage{8959}--\blpage{8971}
(\byear{2020})
\end{barticle}
\endbibitem

\bibitem{janchiv2013time}
\begin{barticle}
\bauthor{\bsnm{Janchiv}, \binits{A.}},
\bauthor{\bsnm{Batsaikhan}, \binits{D.}},
\bauthor{\bsnm{Kim}, \binits{B.}},
\bauthor{\bsnm{Lee}, \binits{W.G.}},
\bauthor{\bsnm{Lee}, \binits{S.-G.}}:
\batitle{Time-efficient and complete coverage path planning based on flow
  networks for multi-robots}.
\bjtitle{International Journal of Control, Automation and Systems}
\bvolume{11}(\bissue{2}),
\bfpage{369}--\blpage{376}
(\byear{2013})
\end{barticle}
\endbibitem

\bibitem{vidal2020concise}
\begin{barticle}
\bauthor{\bsnm{Vidal}, \binits{T.}},
\bauthor{\bsnm{Laporte}, \binits{G.}},
\bauthor{\bsnm{Matl}, \binits{P.}}:
\batitle{A concise guide to existing and emerging vehicle routing problem
  variants}.
\bjtitle{European Journal of Operational Research}
\bvolume{286}(\bissue{2}),
\bfpage{401}--\blpage{416}
(\byear{2020})
\end{barticle}
\endbibitem

\bibitem{eshtehadi2020solving}
\begin{barticle}
\bauthor{\bsnm{Eshtehadi}, \binits{R.}},
\bauthor{\bsnm{Demir}, \binits{E.}},
\bauthor{\bsnm{Huang}, \binits{Y.}}:
\batitle{Solving the vehicle routing problem with multi-compartment vehicles
  for city logistics}.
\bjtitle{Computers \& Operations Research}
\bvolume{115},
\bfpage{104859}
(\byear{2020})
\end{barticle}
\endbibitem

\bibitem{guan2021kohonen}
\begin{bchapter}
\bauthor{\bsnm{Guan}, \binits{Q.}},
\bauthor{\bsnm{Hong}, \binits{X.}},
\bauthor{\bsnm{Ke}, \binits{W.}},
\bauthor{\bsnm{Zhang}, \binits{L.}},
\bauthor{\bsnm{Sun}, \binits{G.}},
\bauthor{\bsnm{Gong}, \binits{Y.}}:
\bctitle{Kohonen self-organizing map based route planning: A revisit}.
In: \bbtitle{2021 IEEE/RSJ International Conference on Intelligent Robots and
  Systems (IROS)},
pp. \bfpage{7969}--\blpage{7976}
(\byear{2021}).
\bcomment{IEEE}
\end{bchapter}
\endbibitem

\bibitem{apuroop2021reinforcement}
\begin{barticle}
\bauthor{\bsnm{Apuroop}, \binits{K.G.S.}},
\bauthor{\bsnm{Le}, \binits{A.V.}},
\bauthor{\bsnm{Elara}, \binits{M.R.}},
\bauthor{\bsnm{Sheu}, \binits{B.J.}}:
\batitle{Reinforcement learning-based complete area coverage path planning for
  a modified htrihex robot}.
\bjtitle{Sensors}
\bvolume{21}(\bissue{4}),
\bfpage{1067}
(\byear{2021})
\end{barticle}
\endbibitem

\bibitem{zhao2022task}
\begin{botherref}
\oauthor{\bsnm{Zhao}, \binits{B.}},
\oauthor{\bsnm{Dong}, \binits{H.}},
\oauthor{\bsnm{Wang}, \binits{Y.}},
\oauthor{\bsnm{Pan}, \binits{T.}}:
A task allocation algorithm based on reinforcement learning in spatio-temporal
  crowdsourcing.
Applied Intelligence,
1--18
(2022)
\end{botherref}
\endbibitem

\bibitem{elmaliach2009multi}
\begin{barticle}
\bauthor{\bsnm{Elmaliach}, \binits{Y.}},
\bauthor{\bsnm{Agmon}, \binits{N.}},
\bauthor{\bsnm{Kaminka}, \binits{G.A.}}:
\batitle{Multi-robot area patrol under frequency constraints}.
\bjtitle{Annals of Mathematics and Artificial Intelligence}
\bvolume{57}(\bissue{3}),
\bfpage{293}--\blpage{320}
(\byear{2009})
\end{barticle}
\endbibitem

\bibitem{kolling2008multi}
\begin{bchapter}
\bauthor{\bsnm{Kolling}, \binits{A.}},
\bauthor{\bsnm{Carpin}, \binits{S.}}:
\bctitle{Multi-robot surveillance: an improved algorithm for the graph-clear
  problem}.
In: \bbtitle{2008 IEEE International Conference on Robotics and Automation},
pp. \bfpage{2360}--\blpage{2365}
(\byear{2008}).
\bcomment{IEEE}
\end{bchapter}
\endbibitem

\bibitem{alitappeh2022multi}
\begin{barticle}
\bauthor{\bsnm{Alitappeh}, \binits{R.J.}},
\bauthor{\bsnm{Jeddisaravi}, \binits{K.}}:
\batitle{Multi-robot exploration in task allocation problem}.
\bjtitle{Applied Intelligence}
\bvolume{52}(\bissue{2}),
\bfpage{2189}--\blpage{2211}
(\byear{2022})
\end{barticle}
\endbibitem

\bibitem{otte2020auctions}
\begin{barticle}
\bauthor{\bsnm{Otte}, \binits{M.}},
\bauthor{\bsnm{Kuhlman}, \binits{M.J.}},
\bauthor{\bsnm{Sofge}, \binits{D.}}:
\batitle{Auctions for multi-robot task allocation in communication limited
  environments}.
\bjtitle{Autonomous Robots}
\bvolume{44}(\bissue{3}),
\bfpage{547}--\blpage{584}
(\byear{2020})
\end{barticle}
\endbibitem

\bibitem{yao2020online}
\begin{barticle}
\bauthor{\bsnm{Yao}, \binits{J.}},
\bauthor{\bsnm{Ansari}, \binits{N.}}:
\batitle{Online task allocation and flying control in fog-aided internet of
  drones}.
\bjtitle{IEEE Transactions on Vehicular Technology}
\bvolume{69}(\bissue{5}),
\bfpage{5562}--\blpage{5569}
(\byear{2020})
\end{barticle}
\endbibitem

\bibitem{banks2020multi}
\begin{bchapter}
\bauthor{\bsnm{Banks}, \binits{C.}},
\bauthor{\bsnm{Wilson}, \binits{S.}},
\bauthor{\bsnm{Coogan}, \binits{S.}},
\bauthor{\bsnm{Egerstedt}, \binits{M.}}:
\bctitle{Multi-agent task allocation using cross-entropy temporal logic
  optimization}.
In: \bbtitle{2020 IEEE International Conference on Robotics and Automation
  (ICRA)},
pp. \bfpage{7712}--\blpage{7718}
(\byear{2020}).
\bcomment{IEEE}
\end{bchapter}
\endbibitem

\bibitem{bai2021distributed}
\begin{botherref}
\oauthor{\bsnm{Bai}, \binits{X.}},
\oauthor{\bsnm{Yan}, \binits{W.}},
\oauthor{\bsnm{Ge}, \binits{S.S.}}:
Distributed task assignment for multiple robots under limited communication
  range.
IEEE Transactions on Systems, Man, and Cybernetics: Systems
(2021)
\end{botherref}
\endbibitem

\bibitem{jin2019dynamic}
\begin{barticle}
\bauthor{\bsnm{Jin}, \binits{L.}},
\bauthor{\bsnm{Li}, \binits{S.}},
\bauthor{\bsnm{La}, \binits{H.M.}},
\bauthor{\bsnm{Zhang}, \binits{X.}},
\bauthor{\bsnm{Hu}, \binits{B.}}:
\batitle{Dynamic task allocation in multi-robot coordination for moving target
  tracking: A distributed approach}.
\bjtitle{Automatica}
\bvolume{100},
\bfpage{75}--\blpage{81}
(\byear{2019})
\end{barticle}
\endbibitem

\bibitem{attiya2020job}
\begin{botherref}
\oauthor{\bsnm{Attiya}, \binits{I.}},
\oauthor{\bsnm{Abd~Elaziz}, \binits{M.}},
\oauthor{\bsnm{Xiong}, \binits{S.}}:
Job scheduling in cloud computing using a modified harris hawks optimization
  and simulated annealing algorithm.
Computational intelligence and neuroscience
\textbf{2020}
(2020)
\end{botherref}
\endbibitem

\bibitem{inceoglu2018continuous}
\begin{barticle}
\bauthor{\bsnm{Inceoglu}, \binits{A.}},
\bauthor{\bsnm{Koc}, \binits{C.}},
\bauthor{\bsnm{Kanat}, \binits{B.O.}},
\bauthor{\bsnm{Ersen}, \binits{M.}},
\bauthor{\bsnm{Sariel}, \binits{S.}}:
\batitle{Continuous visual world modeling for autonomous robot manipulation}.
\bjtitle{IEEE Transactions on Systems, Man, and Cybernetics: Systems}
\bvolume{49}(\bissue{1}),
\bfpage{192}--\blpage{205}
(\byear{2018})
\end{barticle}
\endbibitem

\bibitem{soyster1973convex}
\begin{barticle}
\bauthor{\bsnm{Soyster}, \binits{A.L.}}:
\batitle{Convex programming with set-inclusive constraints and applications to
  inexact linear programming}.
\bjtitle{Operations research}
\bvolume{21}(\bissue{5}),
\bfpage{1154}--\blpage{1157}
(\byear{1973})
\end{barticle}
\endbibitem

\bibitem{mulvey1995robust}
\begin{barticle}
\bauthor{\bsnm{Mulvey}, \binits{J.M.}},
\bauthor{\bsnm{Vanderbei}, \binits{R.J.}},
\bauthor{\bsnm{Zenios}, \binits{S.A.}}:
\batitle{Robust optimization of large-scale systems}.
\bjtitle{Operations research}
\bvolume{43}(\bissue{2}),
\bfpage{264}--\blpage{281}
(\byear{1995})
\end{barticle}
\endbibitem

\bibitem{li2012robust}
\begin{bchapter}
\bauthor{\bsnm{Li}, \binits{Z.}},
\bauthor{\bsnm{Floudas}, \binits{C.A.}}:
\bctitle{Robust counterpart optimization: Uncertainty sets, formulations and
  probabilistic guarantees}.
In: \bbtitle{Proceedings of the 6th Conference on Foundations of Computer-aided
  Process Operations, Savannah (Georgia)}
(\byear{2012})
\end{bchapter}
\endbibitem

\bibitem{ordonez2010robust}
\begin{bchapter}
\bauthor{\bsnm{Ord{\'o}{\~n}ez}, \binits{F.}}:
\bctitle{Robust vehicle routing}.
In: \bbtitle{Risk and Optimization in an Uncertain World},
pp. \bfpage{153}--\blpage{178}.
\bpublisher{INFORMS}, \blocation{???}
(\byear{2010})
\end{bchapter}
\endbibitem

\bibitem{ei1997robust}
\begin{barticle}
\bauthor{\bsnm{EI-Ghaoui}, \binits{L.}},
\bauthor{\bsnm{Lebret}, \binits{H.}}:
\batitle{Robust solutions to least-square problems to uncertain data matrices}.
\bjtitle{Sima Journal on Matrix Analysis and Applications}
\bvolume{18},
\bfpage{1035}--\blpage{1064}
(\byear{1997})
\end{barticle}
\endbibitem

\bibitem{ben1997robust}
\begin{barticle}
\bauthor{\bsnm{Ben-Tal}, \binits{A.}},
\bauthor{\bsnm{Nemirovski}, \binits{A.}}:
\batitle{Robust truss topology design via semidefinite programming}.
\bjtitle{SIAM journal on optimization}
\bvolume{7}(\bissue{4}),
\bfpage{991}--\blpage{1016}
(\byear{1997})
\end{barticle}
\endbibitem

\bibitem{tajik2014robust}
\begin{barticle}
\bauthor{\bsnm{Tajik}, \binits{N.}},
\bauthor{\bsnm{Tavakkoli-Moghaddam}, \binits{R.}},
\bauthor{\bsnm{Vahdani}, \binits{B.}},
\bauthor{\bsnm{Mousavi}, \binits{S.M.}}:
\batitle{A robust optimization approach for pollution routing problem with
  pickup and delivery under uncertainty}.
\bjtitle{Journal of Manufacturing Systems}
\bvolume{33}(\bissue{2}),
\bfpage{277}--\blpage{286}
(\byear{2014})
\end{barticle}
\endbibitem

\bibitem{saeedvand2019robust}
\begin{barticle}
\bauthor{\bsnm{Saeedvand}, \binits{S.}},
\bauthor{\bsnm{Aghdasi}, \binits{H.S.}},
\bauthor{\bsnm{Baltes}, \binits{J.}}:
\batitle{Robust multi-objective multi-humanoid robots task allocation based on
  novel hybrid metaheuristic algorithm}.
\bjtitle{Applied Intelligence}
\bvolume{49},
\bfpage{4097}--\blpage{4127}
(\byear{2019})
\end{barticle}
\endbibitem

\bibitem{ben2006extending}
\begin{barticle}
\bauthor{\bsnm{Ben-Tal}, \binits{A.}},
\bauthor{\bsnm{Boyd}, \binits{S.}},
\bauthor{\bsnm{Nemirovski}, \binits{A.}}:
\batitle{Extending scope of robust optimization: Comprehensive robust
  counterparts of uncertain problems}.
\bjtitle{Mathematical Programming}
\bvolume{107}(\bissue{1}),
\bfpage{63}--\blpage{89}
(\byear{2006})
\end{barticle}
\endbibitem

\bibitem{bertsimas2004price}
\begin{barticle}
\bauthor{\bsnm{Bertsimas}, \binits{D.}},
\bauthor{\bsnm{Sim}, \binits{M.}}:
\batitle{The price of robustness}.
\bjtitle{Operations research}
\bvolume{52}(\bissue{1}),
\bfpage{35}--\blpage{53}
(\byear{2004})
\end{barticle}
\endbibitem

\bibitem{hart1968formal}
\begin{barticle}
\bauthor{\bsnm{Hart}, \binits{P.E.}},
\bauthor{\bsnm{Nilsson}, \binits{N.J.}},
\bauthor{\bsnm{Raphael}, \binits{B.}}:
\batitle{A formal basis for the heuristic determination of minimum cost paths}.
\bjtitle{IEEE transactions on Systems Science and Cybernetics}
\bvolume{4}(\bissue{2}),
\bfpage{100}--\blpage{107}
(\byear{1968})
\end{barticle}
\endbibitem

\bibitem{sungur2008robust}
\begin{barticle}
\bauthor{\bsnm{Sungur}, \binits{I.}},
\bauthor{\bsnm{Ord{\'o}nez}, \binits{F.}},
\bauthor{\bsnm{Dessouky}, \binits{M.}}:
\batitle{A robust optimization approach for the capacitated vehicle routing
  problem with demand uncertainty}.
\bjtitle{Iie Transactions}
\bvolume{40}(\bissue{5}),
\bfpage{509}--\blpage{523}
(\byear{2008})
\end{barticle}
\endbibitem

\bibitem{kirkpatrick1983optimization}
\begin{barticle}
\bauthor{\bsnm{Kirkpatrick}, \binits{S.}},
\bauthor{\bsnm{Gelatt}, \binits{C.D.}},
\bauthor{\bsnm{Vecchi}, \binits{M.P.}}:
\batitle{Optimization by simulated annealing}.
\bjtitle{science}
\bvolume{220}(\bissue{4598}),
\bfpage{671}--\blpage{680}
(\byear{1983})
\end{barticle}
\endbibitem

\bibitem{holland1992genetic}
\begin{barticle}
\bauthor{\bsnm{Holland}, \binits{J.H.}}:
\batitle{Genetic algorithms}.
\bjtitle{Scientific american}
\bvolume{267}(\bissue{1}),
\bfpage{66}--\blpage{73}
(\byear{1992})
\end{barticle}
\endbibitem

\bibitem{bansal2019particle}
\begin{bchapter}
\bauthor{\bsnm{Bansal}, \binits{J.C.}}:
\bctitle{Particle swarm optimization}.
In: \bbtitle{Evolutionary and Swarm Intelligence Algorithms},
pp. \bfpage{11}--\blpage{23}.
\bpublisher{Springer}, \blocation{???}
(\byear{2019})
\end{bchapter}
\endbibitem

\bibitem{liu2015rent3d}
\begin{bchapter}
\bauthor{\bsnm{Liu}, \binits{C.}},
\bauthor{\bsnm{Schwing}, \binits{A.G.}},
\bauthor{\bsnm{Kundu}, \binits{K.}},
\bauthor{\bsnm{Urtasun}, \binits{R.}},
\bauthor{\bsnm{Fidler}, \binits{S.}}:
\bctitle{Rent3d: Floor-plan priors for monocular layout estimation}.
In: \bbtitle{Proceedings of the IEEE Conference on Computer Vision and Pattern
  Recognition},
pp. \bfpage{3413}--\blpage{3421}
(\byear{2015})
\end{bchapter}
\endbibitem

\bibitem{10.5555/3327546.3327651}
\begin{bchapter}
\bauthor{\bsnm{Nazari}, \binits{M.}},
\bauthor{\bsnm{Oroojlooy}, \binits{A.}},
\bauthor{\bsnm{Tak\'{a}\v{c}}, \binits{M.}},
\bauthor{\bsnm{Snyder}, \binits{L.V.}}:
\bctitle{Reinforcement learning for solving the vehicle routing problem}.
In: \bbtitle{Proceedings of the 32nd International Conference on Neural
  Information Processing Systems}.
\bsertitle{NIPS'18},
pp. \bfpage{9861}--\blpage{9871}.
\bpublisher{Curran Associates Inc.},
\blocation{Red Hook, NY, USA}
(\byear{2018})
\end{bchapter}
\endbibitem

\end{thebibliography}

\end{document}